\documentclass{article}

\usepackage[nonatbib,final]{neurips_2022} 

\newif\ifchecklist 
\checklistfalse

\newcommand{\solverExampleFontSize}{\footnotesize}

\usepackage[utf8]{inputenc} 
\usepackage[T1]{fontenc}    
\usepackage{hyperref}       
\usepackage{url}            
\usepackage{booktabs}       
\usepackage{amsfonts}       
\usepackage{nicefrac}       
\usepackage{microtype}      
\usepackage{subcaption}
\usepackage{graphicx}
\usepackage{sidecap}
\usepackage{lipsum}
\usepackage{amsmath}
\usepackage{wrapfig}
\usepackage{paralist}
\usepackage{centernot}
\usepackage{longtable}
\usepackage{pbox}

\usepackage{listings}
\usepackage{tabularx}
\usepackage{xcolor}
\usepackage{wasysym}
\usepackage{microtype}
\usepackage[scaled=0.85]{FiraMono}
\DisableLigatures{encoding = *, family = tt*}

\makeatletter  

\@ifpackagelater{listings}{2018/09/03}
{

\makeatletter
\let\old@lstKV@SwitchCases\lstKV@SwitchCases
\def\lstKV@SwitchCases#1#2#3{}
\makeatother
\usepackage{lstlinebgrd}
\makeatletter
\let\lstKV@SwitchCases\old@lstKV@SwitchCases

\lst@Key{numbers}{none}{%
    \def\lst@PlaceNumber{\lst@linebgrd}%
    \lstKV@SwitchCases{#1}%
    {none:\\%
     left:\def\lst@PlaceNumber{\llap{\normalfont
                \lst@numberstyle{\thelstnumber}\kern\lst@numbersep}\lst@linebgrd}\\%
     right:\def\lst@PlaceNumber{\rlap{\normalfont
                \kern\linewidth \kern\lst@numbersep
                \lst@numberstyle{\thelstnumber}}\lst@linebgrd}%
    }{\PackageError{Listings}{Numbers #1 unknown}\@ehc}}
\makeatother

}
{

\usepackage{lstlinebgrd}

}

\makeatother
\usepackage{pgffor}
\makeatletter

\newcount\bt@rangea
\newcount\bt@rangeb

\newcommand\btIfInRange[2]{%
    \global\let\bt@inrange\@secondoftwo%
    \edef\bt@rangelist{#2}%
    \foreach \range in \bt@rangelist {%
        \afterassignment\bt@getrangeb%
        \bt@rangea=0\range\relax%
        \pgfmathtruncatemacro\result{ ( #1 >= \bt@rangea) && (#1 <= \bt@rangeb) }%
        \ifnum\result=1\relax%
            \breakforeach%
            \global\let\bt@inrange\@firstoftwo%
        \fi%
    }%
    \bt@inrange%
}
\newcommand\bt@getrangeb{%
    \@ifnextchar\relax%
        {\bt@rangeb=\bt@rangea}%
        {\@getrangeb}%
}
\def\@getrangeb-#1\relax{%
    \ifx\relax#1\relax%
        \bt@rangeb=100000
    \else%
        \bt@rangeb=#1\relax%
    \fi%
}

\newcommand{\btLstHL}[2]{
  \btIfInRange{\value{lstnumber}}{#1}{#2}{}
}

\makeatother

\newcommand{\lstleftmargin}{0.8cm}

\newcommand{\setlstc}{
  \lstset{
    basicstyle=\normalsize\ttfamily,
    columns=fullflexible,
    keywords=[2]{while, assert, assume},
  }
}

\newcommand{\setlstpy}{
  \lstset{
    language=Python,
    numbersep=10pt,  
    basicstyle=\scriptsize\ttfamily,
    numberstyle=\scriptsize\ttfamily,
    columns=fullflexible,
    morekeywords={case, match, nonlocal, assert, reward, event},
    deletekeywords=[2]{vars, not, min, max, len, bool},
    deletekeywords=[1]{not},
    emph={choose, checkpoint},
    emphstyle=\ttfamily\textbf,
    stringstyle=\color{brown!80!black},
    showstringspaces=false
  }
}

\makeatletter\let\expandableinput\@@input\makeatother

\newcommand{\looprl}{Looprl}
\newcommand{\limply}{\ \rightarrow \ }

\newcommand{\pfail}{p_{0}}
\newcommand{\pmin}{p_{1}}
\newcommand{\psucc}{p_{2}}
\newcommand{\rmin}{r_{\texttt{min}}}

\newcommand{\myparagraph}[1]{\noindent\textbf{#1} \ }

\newcommand{\subtitlestyle}[1]{{\large #1}}

\newcommand{\subtitlesep}{\\[0.1em]}

\newcommand{\subtitle}[1]{\subtitlesep{} \subtitlestyle{#1}}

\title{{Learning to Find Proofs and Theorems by \\ Learning to Refine Search Strategies} \subtitle{The Case of Loop Invariant Synthesis}}

\author{
  Jonathan Laurent \\
  Carnegie Mellon University \\
  Karlsruhe Institute of Technology \\
\And
  André Platzer \\
  Carnegie Mellon University \\
  Karlsruhe Institute of Technology \\
}

\begin{document}

\maketitle

\begin{abstract}
  We propose a new approach to automated theorem proving where an AlphaZero-style agent is self-training to refine a generic high-level expert strategy expressed as a nondeterministic program. An analogous teacher agent is self-training to generate tasks of suitable relevance and difficulty for the learner. This allows leveraging minimal amounts of domain knowledge to tackle problems for which training data is unavailable or hard to synthesize. As a specific illustration, we consider loop invariant synthesis for imperative programs and use neural networks to refine both the teacher and solver strategies.
\end{abstract}


\section{Introduction}

Augmenting tactic-based interactive theorem provers with neural guidance has been the focus of increased attention in recent years \cite{DBLP:conf/iclr/HuangDSS19,DBLP:conf/icml/YangD19,DBLP:conf/icml/BansalLRSW19,DBLP:conf/iclr/LiYWP21,DBLP:conf/iclr/HanRWAP22}. The dominant approach uses imitation learning on corpora of formalized mathematics. However, despite recent efforts involving self-supervised pre-training \cite{DBLP:conf/iclr/HanRWAP22} or data-augmentation \cite{DBLP:conf/nips/WangD20}, this approach is limited by the conspicuous scarcity of human-produced training data. An alternative approach inspired by the success of AlphaZero \cite{silver2018general} is to use reinforcement learning and let an agent self-train to interact with a theorem prover via trial and error \cite{DBLP:journals/corr/abs-1905-10501,DBLP:conf/nips/WuNWD21}. However, previous attempts of doing so have been hampered by two fundamental issues:

\begin{itemize}
\item First, existing tactic-based theorem provers offer infinite action spaces not amenable to random exploration. Also, they are optimized for formalizing the outcome of human insights concisely but often fail to define a tractable search space for deriving those insights in the first place \cite{selsambeyond}. For example, using tactics effectively often requires providing insights in advance that are more easily found later in the search process (e.g. constructing a term to instantiate an existential quantifier \cite{DBLP:journals/jar/HahnleS94}). More generally, most of the human process of discovering proofs, which involves trial and error, sketching, and abductive reasoning is not captured by standard prover tactics and so we can hardly expect a reinforcement learning agent to learn this process through sheer interaction with a tactic-based prover.
\item Second, although reinforcement learning alleviates the need for human proofs during training, the agent must still be provided learning tasks of suitable relevance, diversity, difficulty, and generalizability in the form of theorem statements to be proved. This is in contrast with applications of AlphaZero in board games, where the symmetric nature of games such as Chess or Go enables leveraging symmetric self-play as an infinite source of training tasks.
\end{itemize}



In this paper, we suggest a novel approach to learning automated theorem proving that does not rely on human-provided proofs and theorems. Instead, a teacher agent is trained to generate interesting and relevant tasks while a solver agent is co-trained to solve them. Our core insight is that \emph{both} agents can be implemented by using reinforcement learning to refine high-level search \emph{strategies}~\cite{selsambeyond} written by experts in the form of nondeterministic programs. Our paper also defines novel design principles and language constructs for expressing such strategies. These include \emph{conditional generative strategies} as a template for implementing teachers, \emph{abductive reasoning} as a design principle that is made scalable by self-learned guidance, and \emph{strategy events} as an abstraction that enables easier reward engineering and better sample-efficiency. This paper introduces our new framework and also evaluates it in a well-contained yet challenging setting, namely the automatic verification of imperative programs with loops. We frame our contributions in a way to emphasize general applicability. An empirical evaluation in other application domains is left to future work.


\section{Approach}


\subsection{Background: Verifying Imperative Programs with Loops}

Suppose we are given a program such as the one in Figure~\ref{fig:example-prog} and we want to prove that the final assertion always holds. The way to proceed is to find a predicate called a \emph{loop invariant} with the following properties:
\begin{inparaenum}[\it i)]
  \item it is true before the loop is entered,
  \item it is preserved by the loop body when the loop guard holds and
  \item it implies the final assertion when the loop guard does not hold.
\end{inparaenum}
By ``preserved'', we mean that if the invariant is true before executing the loop body then it is also true afterwards.
Finding loop invariants is the most crucial aspect of program verification \cite{DBLP:journals/siamcomp/Cook78,DBLP:books/sp/Harel79} and still resists automation \cite{DBLP:conf/cade/Filliatre13}.

\begin{figure}
  \begin{center}
    \begin{minipage}{0.28\textwidth}
      \begin{minipage}{0.20\textwidth}
        \input{content/example-prog.tex}
      \end{minipage}
    \end{minipage}
    \begin{minipage}{0.70\textwidth}
      A desirable \emph{invariant} is a formula $I[x, y]$ such that:
      \begin{equation*}
        \forall x, y \
        \begin{cases}
          x \geq 1 \,\wedge\, y = 0 \limply I[x,y] & \ \text{(holds initially)} \\
          y < 1000 \,\wedge\, I[x, y] \limply I[x+y, y+1] & \ \text{(preserved)} \\
          y \geq 1000 \,\wedge\, I[x, y] \limply x \ge y & \ \text{(implies post)}
        \end{cases}
      \end{equation*}

      Such an invariant is $I[x, y] := (x \geq y \,\wedge\,  x \geq 1 \,\wedge\, y \geq 0)$
    \end{minipage}
  \end{center}
  \hrule
  \medskip
  \caption{An example program and an associated loop invariant}\label{fig:example-prog}
\end{figure}

Despite the difficulty of automatically synthesizing loop invariants, humans do so routinely using well-understood search strategies. For example, in the case of the example in Figure~\ref{fig:example-prog}, one would first try to prove the postcondition $x \geq y$ itself as an invariant and then note that it is not preserved by the loop body as the following implication does not hold: $y < 1000 \,\wedge\, x \geq y \centernot\limply x + y \geq y + 1$. However, simplifying the right-hand side, we see that the proof works if we can show $x \geq 1$ to be an invariant itself. Using a similar form of abductive reasoning, we find that this in turn requires $y \geq 0$ to be an invariant and we end up proposing $x \geq y \wedge x \geq 1 \wedge y \geq 0$ as a loop invariant.

\subsection{Expressing Search Strategies as Nondeterministic Programs}

The search strategy we followed for the example above can be formalized as a nondeterministic program, which is shown in Figure~\ref{fig:simplified-solver}. As a nondeterministic program, it features a \texttt{choose} operator that takes a list of objects as an input and selects one of them nondeterministically. A nondeterministic program can be refined into a deterministic one by providing an external oracle to implement the \texttt{choose} operator. It also defines a search tree that can be explored with or without a guiding heuristic. In this work, we use neural networks to implement such oracles and guiding heuristics.

\begin{figure}
  \setlstpy{}
  \centering
    \begin{minipage}{0.75\textwidth}
      \input{content/simplified-solver.tex}
    \end{minipage}
  \caption{A simple strategy for finding loop invariants, described in Python-like pseudocode syntax. In this strategy, an initial invariant candidate is selected nondeterministically that implies the postcondition (line~\ref{line:prove-post}). One ensures that this candidate holds initially (line~\ref{line:prove-init}), without which the strategy fails immediately. Then, one attempts to prove that it is preserved by the loop body (lines~\ref{line:ind-obl} and \ref{line:ind-obl-abduct}, where \texttt{wlp} denotes Hoare's weakest liberal precondition operator \cite{DBLP:journals/cacm/Hoare69}). If the candidate is not preserved, the \texttt{abduct} function suggests a list of assumptions that make it so. One then uses the \texttt{choose} operator to nondeterministically select one of them, which we try and prove invariant recursively (line~\ref{line:prove-inv-rec}). Because the number of abduction candidates to choose from can be large, successfully using such a strategy depends on having an effective oracle to guide search. To provide sufficient context to such an oracle, calls to \texttt{choose} must provide some extra information that is omitted here for brevity. In this case, we would pass a special token to indicate the call site, the program being analyzed and the value of \texttt{inv} in the case of line~\ref{line:choose-inv}. See Appendix~\ref{ap:solver-details} for more details and for a full listing of the invariant synthesis strategy that we use in our experiments. 
  }\label{fig:simplified-solver}
\end{figure}


In addition, a key aspect of this strategy is to infer missing assumptions under which a currently failing proof obligation would hold. This form of reasoning is called \emph{abductive reasoning} and it is fundamental in the way humans search for proofs~\cite{meyer2010abduction,polya2004solve}. It is implemented in a separate \texttt{abduct} procedure that takes a formula as an input and returns either \texttt{Valid} or a list of abduction candidates. For example, the \texttt{abduct} procedure fails to prove the implication $x \ge 0 \rightarrow x + y \ge 1$ but may suggest $x + y \ge 1$, $x < 0$ and $y \ge 1$ as possible missing assumptions. The use of abductive reasoning for theorem proving and loop invariant synthesis specifically has been proposed in the past \cite{DBLP:conf/oopsla/DilligDLM13,DBLP:conf/frocos/EchenimPS19}. However, abduction procedures are hard to implement~\cite{DBLP:journals/jacm/EiterG95} and typically only available for specific decidable theories. Also, using them in proof search tends to scale poorly in the absence of good heuristics to rank and filter abduction candidates. Our proposed framework makes abductive reasoning practical by addressing both issues: it allows leveraging self-learned guidance to select abduction candidates and it allows implementing abduction procedures as nondeterministic programs that are amenable to learning themselves.


Another notable feature of the strategy in Figure~\ref{fig:simplified-solver} is the use of the \texttt{reward} operator on line~\ref{line:reward-ex} to incentivize finding short invariants (short proofs are desirable in general as they tend to be easier to check, interpret and generalize). The \texttt{reward} operator can be used multiple times and at any point of the strategy execution. An implicit reward of 1 or -1 is emitted when the strategy successfully returns or fails respectively. An optimal execution of a nondeterministic strategy is one that maximizes the cumulative amount of collected rewards. In this example, we bound the maximal invariant size penalty in such a way to ensure that finding a proof is always rewarded more than failing at doing so.

As argued by Selsam \cite{selsambeyond}, defining search strategies as nondeterministic programs provides a natural and flexible way for experts to leverage domain-specific knowledge while outsourcing difficult proof decisions to search algorithms and learned heuristics. This paradigm differs from the standard paradigm of tactic-based theorem provers in which an external entity must orchestrate stateless tactics that do not call for any form of interactive guidance internally. In contrast, the nondeterministic programming approach \emph{inverts} control and has strategies call external oracles rather than the other way around, which allows for a much tighter control of the resulting search space.

\subsection{Refining Nondeterministic Strategies with Reinforcement Learning}

A nondeterministic program induces a (deterministic) Markov Decision Process (MDP) where intermediate states are choice points and final states are either success states (the program returns successfully) or failure states (an assertion is violated or \texttt{choose} is called on an empty list of choices). Standard search algorithms can be used to navigate this MDP but doing so in an efficient and scalable way requires strong heuristics for guiding search.

In this work, we propose to learn such heuristics in a self-supervised fashion using the AlphaZero algorithm \cite{silver2018general,DBLP:conf/iclr/DanihelkaGSS22}. In AlphaZero, a neural network is trained to map any state to a probability distribution over available actions along with a value estimate (i.e. an estimate of the expected sum of future rewards).
The neural network alone can be used as a policy for navigating the MDP. However, a stronger policy results from combining the network with a tree search algorithm such as Monte-Carlo Tree Search (MCTS) \cite{DBLP:journals/tciaig/BrownePWLCRTPSC12}. The key insight underlying AlphaZero is that the network can be trained via an iterative improvement process where it is successively
\begin{inparaenum}[\it i)]
\item used as an MCTS heuristic to try and solve problems and then
\item updated using gradient descent so as to better predict the outcome of each attempt along with the action selected by MCTS on all states encountered along the way.
\end{inparaenum}
More details on AlphaZero can be found in the literature \cite{silver2018general}.

Using AlphaZero, we can refine nondeterministic proof search strategies \emph{without} external proof examples to learn from. However, AlphaZero must still be provided with a set of training problems to be solved. Training problems should be diverse, relevant and numerous enough to allow proper generalization. They should also be of varied difficulty to allow for learning to bootstrap.

\subsection{Generating Training Problems with Conditional Generative Strategies}

In theorem proving, high-quality training problems produced by humans are often not available in quantities even remotely matching the needs of reinforcement learning to properly generalize across problem instances. A possible solution is to use procedural generation techniques to assemble large datasets of synthetic problems. This works well in some specific areas \cite{DBLP:conf/iclr/BunelHDSK18} and particularly for problems that can be framed as inverse problems such as symbolic integration~\cite{DBLP:conf/iclr/LampleC20}. In other areas, generating interesting problems is as hard as solving existing ones, possibly harder.

This is the case in particular with the problem of loop invariant synthesis. Here, a natural idea for generating problem instances would be to use a probabilistic grammar for repeatedly sampling triples consisting of a program, a loop invariant and an assertion and then reject all triples in which the properties defining valid invariants do not hold. Unfortunately, not only would doing so naively lead to a very high rejection rate, but the resulting dataset would be heavily biased towards trivial samples that are hardest to reject (e.g. programs where the final assertion is the negation of the loop guard and where the invariant does not even matter). In contrast, many classes of interesting problems from standard human-written benchmarks would only be sampled with an infinitesimal probability.

In this work, we propose to generate problems using the same methodology we use to solve them: by leveraging reinforcement learning to refine nondeterministic search strategies.

Specifically, experts can define \emph{teacher strategies} in the form of stochastic and nondeterministic programs, each run of which either fails or successfully returns with a problem instance. Reinforcement learning can then be used to refine such programs with the two objectives of maximizing the diversity and interestingness of generated problems while avoiding rejection. However, these objectives are naturally in tension since an agent may be locally incentivized to avoid generating particular types of problems that are easier to reject. We introduce a design template for a class of strategies that avoid this obstacle. We call such strategies \emph{conditional generative strategies}.

A conditional generative strategy generates a problem in three steps:
\begin{inparaenum}[\it i)]
\item it samples a set of constraints that define desired features for the generated problem,
\item it nondeterministically generates a problem that respects as many constraints as possible and gets rewarded for doing so and
\item it applies a sequence of random and validity-preserving problem transformations as a way to further increase diversity.
\end{inparaenum}
We provide a high-level description of a such a teacher strategy for invariant synthesis in Figure~\ref{fig:teacher-simple}.
 The interest of such an architecture is that it enables decoupling the two conflicting objectives of generating valid and diverse problems: diversity is guaranteed by the first and third steps while learning can focus on the second step. This also allows using the exact same learning algorithm for training both teacher and solver agents (AlphaZero in this work).

\begin{figure}
  \setlstpy{}
  \begin{center}
    \noindent\begin{minipage}{1.0\textwidth}
      \input{content/simplified-teacher.tex}
    \end{minipage}
  \end{center}
  \medskip
  \caption{Simplified teacher strategy. Problems are generated by nondeterministically refining the template in lines \ref{line:template:start}-\ref{line:template:end} in a way to optimize random constraints (see examples in Table~\ref{tab:teacher-constrs-subset}). In this template, the invariant is expressed as a conjunction of at most three sub-invariants: \texttt{inv\_lin} is a linear equality or inequality, \texttt{inv\_main} is a disjunction of atomic formulas (i.e. comparisons) and \texttt{inv\_aux} is a conjunction of atomic formulas that can be used in proving \texttt{inv\_main} but not \texttt{post}. These three formulas are refined nondeterministically by the call to \texttt{refine\_inv} on line~\ref{line:refine-inv}. After an invariant is chosen, a loop body is also generated nondeterministically (line~\ref{line:refine-body}) that preserves the invariant (line~\ref{line:inv-preserved}). Finally, the \texttt{num\_violations} function penalizes the presence of useless or redundant problem components. For example, a penalty is applied if \texttt{inv\_main} features a disjunct that can be removed without invalidating the problem. Appendix~\ref{ap:teacher-details} provides more details on how the \texttt{refine\_*} functions can be implemented. The simplest way to do so is to have a grammar and use the \texttt{choose} operator to select rules recursively. However, fancier strategies can easily be implemented in the nondeterministic programming framework. In particular, our implementation uses the \texttt{abduct} procedure introduced in Figure~\ref{fig:simplified-solver} to suggest values for constants and subformulas.}\label{fig:teacher-simple}
\end{figure}

\begin{table}
  \renewcommand{\arraystretch}{1.5}
  \centering
\small
\newcommand{\boolvals}{\texttt{bool}}
\newcommand{\litvals}[1]{\texttt{#1}}
\newcommand{\tsep}{\hline}
\renewcommand\tabularxcolumn[1]{m{#1}} 
\begin{tabularx}{\textwidth}{m{4.3cm}m{1.5cm}X}
\textbf{Name} & \textbf{Type} & \textbf{Description} \\
\toprule
\texttt{num-inv-main-disjuncts} & \litvals{none|1|2} &
If an integer $n$, then \texttt{inv\_main} is refined with a disjunction of $n$ atomic formulas.
\\ \tsep
\texttt{has-conditional} & \boolvals{} &
Whether \texttt{body} must include a conditional statement.
\\ \tsep
\texttt{loop-guard-useful-for-post} & \boolvals{} &
Whether assuming the negation of the loop guard is useful in proving the postcondition \texttt{post}.
\\ \tsep
\texttt{body-implies-main-inv} & \boolvals{} &
If \texttt{true}, then \texttt{inv\_main} always holds after executing \texttt{body} regardless of whether or not it holds before.
\\ \tsep
\texttt{eq-only-for-init} & \boolvals{} &
Whether or not to use equalities only in \texttt{init}.
\\ \bottomrule \\
\end{tabularx}
  \bigskip
  \caption{Examples of teacher constraints. Some descriptions refer to names introduced in Figure~\ref{fig:teacher-simple}. The full list of constraints we used in our experiments is available in Appendix~\ref{ap:teacher-details}. }\label{tab:teacher-constrs-subset}
\end{table}

\subsection{Predicting Events Rather Than Rewards}\label{sec:vec-rewards}

In the standard AlphaZero setting, the network is trained to predict the value of encountered states, that is the expected sum of future rewards. However, such a number conflates a lot of valuable information. For example, suppose a teacher state is assigned a value of 0. Does 0 mean that the teacher is predicted to either fail or succeed with equal probability or that it will certainly succeed in generating a problem that violates two constraints? And if so, which two?

Although this information is not relevant to MCTS, there are several reasons for having the network predict it anyway. First, doing so can result in an increased sample efficiency by providing the network with more detailed feedback on its mistakes. Such an effect has been previously demonstrated in the KataGo project \cite{DBLP:journals/corr/abs-1902-10565}. Second, it is interesting from an interpretability point of view and makes it easier to diagnose problems and weaknesses in a given network. We found a third reason that is more subtle, which has to do with getting the combined benefits of issuing intermediate and final rewards.

Indeed, in the teacher strategy showed in Figure~\ref{fig:teacher-simple}, we only penalize the agent for violating a constraint \emph{at the very end}. Intuitively, there is a lost opportunity here since many violations could be detected much earlier while the problem is still being refined. Emitting an intermediate reward at this point would certainly improve learning efficiency. However, the same violation can be potentially detected at several different points in time and we must ensure that a reward is only issued the first time. This in turns makes the job of the value prediction network harder since it needs to keep track of whether or not each violation has been penalized already. Encoding this information alone can be costly, and especially so for attention-based network architectures such as transformers with a quadratic inference cost. Having the network predict  constraint violations separately offers an elegant way out: all rewards are issued at the end but from the moment a constraint violation is detected, the network's predicted probability for this violation is overridden to 1 whenever a value estimate is computed.

More formally, we propose to introduce the concept of an \emph{event}. Strategies can declare an arbitrary number of events $e$ and each one is associated with a reward $r_e$ along with a maximal number of occurrences $m_e$ ($m_e=1$ for constraint violation events) that can be counted towards the final reward. The \texttt{reward} operator introduced in Figure~\ref{fig:simplified-solver} is replaced by an \texttt{event} operator. When a strategy terminates, a reward is issued implicitly with a value of $-1$ in case of a failure and \[\max\Big\{1 + \sum_e r_e\min(n_e,m_e), \ \rmin\Big\}\] in case of a success. In this expression, $n_e$ denotes the number of calls made to $\texttt{event}(e)$ during the whole episode and setting $\rmin > -1$ guarantees that successes are always rewarded more than failures. Note that it is important not to issue event penalties after a failure. Otherwise, faced with a likely failure, an agent may be incentivized to simply give up and fail immediately to avoid further penalties in search of success. Also, there are two reasons for bounding the maximal number of occurrences of event $e$ that are counted towards the final reward by a constant quantity $m_e$. First, this allows having a network with a finite softmax head predict the number of occurrences of $e$ (as we see below). Second, the $m_e=1$ case is particularly useful to model events such as constraint violations that can be detected multiple times but should only be penalized once.

In this framework, the value head of the network is tasked with predicting the following probabilities:
\begin{inparaenum}[\it i)]
  \item the probability $\pfail$ of a failure,
  \item the probability $\pmin$ of succeeding with minimal reward $\rmin$,
  \item the probability $\psucc = 1 - \pfail - \pmin$ of succeeding with a greater reward and
  \item the probabilities $p_e^i$ that $\min(n_e,m_e)=i$ for all events $e$ and $0 \le i \le m_e$.
\end{inparaenum}
A value estimate derives from these probabilities:
\[ -\pfail + \pmin \cdot \rmin + \psucc \cdot \sum_e \sum_i \hat p_e^i \cdot i \cdot r_e \]
where $\hat p_e^i \propto \mathbf{1}{\{n_e \le i\}} \cdot p_e^i$ is a corrected probability estimate accounting for the number of times $n_e$ that $\texttt{event}(e)$ was raised already. In our invariant synthesis experiments, we use events in both the teacher and solver strategies to represent constraint violations $(m_e=1)$ and encourage short proofs by penalizing individual proof steps respectively $(m_e > 1)$.


\section{Implementation and Engineering Contributions}

\begin{figure}
  \centering
  \includegraphics[width=\textwidth]{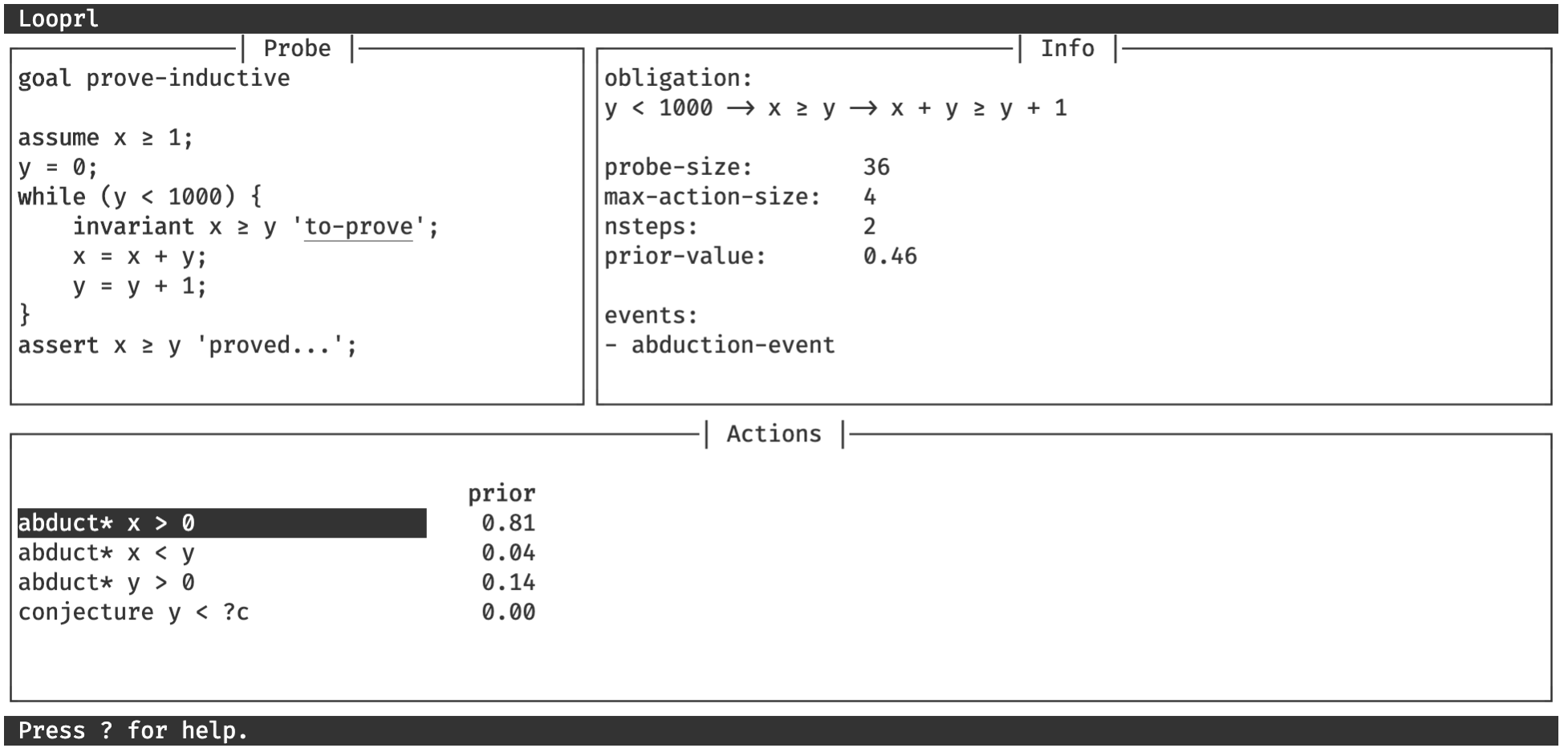}
  \caption{Using the \looprl{} UI to inspect the solver agent on the problem from Figure~\ref{fig:example-prog}. Additional commented screenshots are available in Appendix~\ref{ap:screenshots}.}\label{fig:ui-screenshot}
\end{figure}

This paper advocates for a hybrid approach to automated theorem proving where human experts from a variety of areas can formalize their knowledge succinctly in the form of nondeterministic proving and teaching strategies. Reinforcement learning is leveraged to fill in the blanks in all inevitable cases where the human intuition escapes formalization.

However, for this vision to be practical, writing strategies and iterating on them must be as frictionless as possible. New tools and abstractions are needed to automate the process of converting nondeterministic programs into reinforcement learning environments, ease debugging and foster code reuse. In this work, we accomplish a crucial step in this direction by introducing the \looprl{} theorem prover. \looprl{} consists of the following collection of features:

\begin{itemize}
\item A domain specific language for writing strategies embedded in OCaml,  with first-class support for neural-guided nondeterministic programming. Strategies are automatically compiled into reinforcement learning environments that can be explored in Python.
\item A graphical interface that can be used to interact with strategies manually while inspecting the neural network's predictions and the behavior of MCTS (see Figure~\ref{fig:ui-screenshot}).
\item A library of utility functions for writing program verification strategies. This library includes an implementation of the \texttt{abduct} function introduced in Figure~\ref{fig:simplified-solver} for integer linear arithmetic (see details in Appendix~\ref{ap:abduction}).
\item A parallel, high-performance implementation of the (Gumbel) AlphaZero algorithm~\cite{silver2018general,DBLP:conf/iclr/DanihelkaGSS22} for refining search strategies. Our implementation uses Pytorch~\cite{DBLP:conf/nips/PaszkeGMLBCKLGA19} and Ray~\cite{DBLP:conf/osdi/MoritzNWTLLEYPJ18}. It provides support for events (see Section~\ref{sec:vec-rewards}) and unbounded, variable-size action spaces.
\end{itemize}


The first point on developing a domain-specific language for writing strategies is particularly important and deserves more discussion. In our experience, having such a language is not just a mere matter of convenience but one of feasibility. In fact, we did try to implement an initial version of a teacher strategy for loop invariant synthesis by writing pseudocode on paper and manually compiling it into an MDP implemented in Python. However, doing so resulted in a complexity explosion where any single-line change to the pseudocode could take days of work to implement.

The reason compiling a nondeterministic program into an MDP manually is so tedious is that the whole program state must be made explicit, which includes the program stack. A tempting alternative would be to write nondeterministic code as normal Python code parametrized by an arbitrary \texttt{choose} function. However, this does not allow using tree search on the resulting program since doing so would require cloning the whole execution state of a Python program. Fortunately, there exists a solution to this problem in the field of programming languages using \emph{search monads} \cite{DBLP:journals/corr/Hedges14a,selsambeyond} and which essentially enables writing arbitrary nondeterministic code and then reifying it into a search tree that can be manipulated explicitly.

In \looprl{}, we implement a domain specific language embedded in OCaml for expressing nondeterministic strategies. OCaml is a natural choice because
it is fast (about two orders of magnitude faster than Python for symbolic code such as proof inferences) and it has good support for monads and Python interoperability. Our language provides built-in support for the \texttt{choose} and \texttt{event} operators. Also, it defines an intermediate graph-based format for representing data to be sent to neural networks in an architecture-agnostic way. Any piece of information that is passed to the \texttt{choose} operator must be convertible to this format and feature suitable metadata ensuring a seamless integration with the \looprl{} user interface and debugging tools.


\section{Experiments}\label{sec:experiments}

We tested \looprl{} on the problem of synthesizing loop invariants for single-loop imperative programs and evaluated it on the standard Code2Inv \cite{DBLP:conf/nips/SiDRNS18} benchmark. This benchmark contains a set of 132 programs written in C. All programs feature a single loop along with a final assertion to be proven. They feature linear integer arithmetic and sometimes involve conditionals, assumptions and nondeterministic tests (some Code2Inv problems are in Appendix~\ref{ap:code2inv-examples}). Most existing invariant generation tools can only solve a subset of these problems~\cite{DBLP:conf/cav/SiNDNS20}. To the best of our knowledge, only one existing tool can solve them all~\cite{DBLP:conf/iclr/RyanWYGJ20}. However, it does not use learning and relies on specific optimization techniques that are not generalizable beyond the setting of purely numerical programs.


We used \looprl{} to implement an invariant generation strategy, which we describe in Appendix~\ref{ap:solver-details}. In a nutshell, it generalizes the simple strategy described in Figure~\ref{fig:simplified-solver} by adding features such as the ability to abduct disjunctive invariants or to conjecture invariant templates with placeholder constants to be refined later using abduction. To our surprise and although this generic strategy can be described in only one page of pseudocode, it is sufficient to simply solve \emph{every} Code2Inv benchmark problem in a few seconds when combined with the vanilla MCTS algorithm \cite{DBLP:journals/tciaig/BrownePWLCRTPSC12} (with no learned heuristic). 

We therefore evaluate a self-trained solver agent on its ability to solve as many Code2Inv problems as possible \emph{without} resorting to any search. At every decision point, it greedily takes the highest ranked action according to the network's policy head and no backtracking is allowed. We argue that such a metric is of particular interest and significance. Indeed, finding an invariant for a single loop is of limited interest of its own. Rather, doing so is typically useful as a subtask in a hierarchy of increasingly complex and relevant problems. The next level of this hierarchy may be to prove a piece of code with nested loops, and then to synthesize a function respecting a specification, which is still several levels away from writing full-fledged software systems. Complex problems cannot be solved if backtracking search is needed at every level of this deep hierarchy.


\subsection{Training Protocol}\label{sec:protocol}

We train a teacher agent and a solver agent in sequence using the (Gumbel) AlphaZero algorithm. Both agents use a similar 1.6M parameters Dynamic Graph Transformer network \cite{selsam2020universal} (see architecture details in Appendix~\ref{ap:network}). The teacher agent is trained first for 20 AlphaZero iterations. During each iteration, it goes through 8000 problem generation episodes, using 64 MCTS simulations per step. Then, the network is updated using samples from the $k$ previous iterations with $k$ increasing from 1 to 6 during training. Each iteration simulates 800 additional episodes to generate \emph{validation} samples that are not used to train the network but to control overfitting via early stopping. After the teacher agent is trained, a dataset of 50K problems is gathered for training the solver, which includes a mix of 40K problems generated during the last five training iterations and 10K new ones that are generated without Gumbel exploration noise~\cite{DBLP:conf/iclr/DanihelkaGSS22}. The solver agent is also trained for 20 iterations. During each iteration, it attempts to solve 20K randomly selected problems using 32 MCTS simulations per step. In addition, 5K additional problems are solved to generate validation samples. See Appendix~\ref{ap:hyperparams} for a complete list and explanation of all training hyperparameters.

A complete training run takes about 16h on a single machine with a 10-core Intel i9-10900KF processor, 64GB of RAM and a NVIDIA GeForce RTX 3080 GPU. 

\subsection{Experimental Results}

\newcommand{\nruns}{2}

\begin{figure}
  \centering
  \begin{minipage}{0.45\textwidth}
      \centering
      \includegraphics[width=\textwidth]{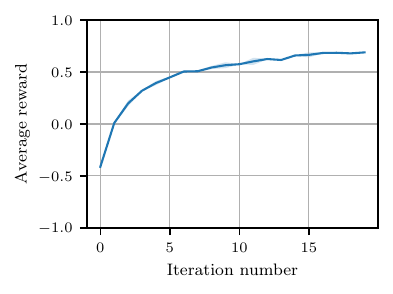}
      \caption{Teacher training curve}\label{fig:teacher-training}
  \end{minipage}\hfill
  \begin{minipage}{0.45\textwidth}
      \centering
      \includegraphics[width=\textwidth]{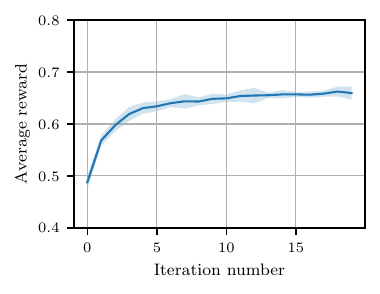}
      \caption{Solver training curve}\label{fig:solver-training}
  \end{minipage}
\end{figure}

We show training curves for the solver and teacher agents in Figures~\ref{fig:teacher-training} and~\ref{fig:solver-training} respectively. These curves record the evolution of the average reward collected by MCTS combined with the latest network during each iteration. Error intervals are estimated based on two random seeds. Note that the maximum theoretical reward obtainable by the teacher agent is ${<}1$ since problem generation constraints may be sampled that are mutually incompatible, in which case the agent simply does its best to minimize the number of violated constraints. The same holds for the solver agent since all generated invariants are penalized proportionally to their size. The solver agent goes through a more modest reward gain during training (the $y$-axis is rescaled in Figure~\ref{fig:solver-training} to emphasize the trend). This is reflective of the fact that most problems generated by the teacher can be solved by MCTS alone and so most of the training concentrates on learning to solve about 10\% of hard problems.

We show the results of our evaluation on the Code2Inv benchmark in Table~\ref{tab:code2inv}. Three different agents are compared on the average ratio of benchmark problems for which they can successfully generate an invariant of minimal size \emph{without} search or backtracking. The first agent is a baseline that simply selects proof actions at random. The second agent is a solver network that was trained using problems generated by an untrained teacher (i.e. using MCTS but no network heuristic). Finally, the third agent is a solver network trained using the full protocol of Section~\ref{sec:protocol}. These results demonstrate the critical impact of learning for both the teacher and solver agents. Unsurprisingly, an inferior teacher leads to an inferior solver with decreased generalization capabilities.



\begin{table}
  \centering
  \begin{tabular}{lc}
  \toprule
  Policy & \% Problems solved \\
  \midrule
  Random & ${18.4 \ \pm\  0.0}$ \\
  Network (untrained teacher) & ${39.7 \ \pm\ 1.6}$ \\
  Network (trained teacher) & $\mathbf{61.5 \ \pm\  0.4}$ \\
  \bottomrule
\end{tabular}
  \bigskip
  \caption{Experimental results on the Code2Inv benchmark (using no backtracking search). The score for the Random heuristic is computed as an average across 10K attempts. Standard deviations are computed based on using two different random seeds for the full training experiment.}\label{tab:code2inv}
\end{table}



\section{Related Work}

\myparagraph{Leveraging nondeterministic programming for proof search.} The idea of combining nondeterministic expert strategies with neural oracles was proposed by Selsam~\cite{selsambeyond,selsamAITPTalk} as a possible angle for tackling the IMO Grand Challenge~\cite{selsamStrategyML}. However, how to train such oracles remains an open problem. One proposal~\cite{selsam2020universal} is to use supervised learning for training a universal oracle that is conditioned on the description of arbitrary search problems and can therefore use training data from a large number of heterogeneous sources. In contrast, we propose using reinforcement learning with the insight that nondeterministic search strategies can be used to \emph{generate} problems in addition to solving them.

\myparagraph{Learning to generate synthetic theorems.} Procedural generation techniques have been used for producing synthetic theorems \cite{DBLP:conf/iclr/WuJBG21,DBLP:journals/corr/abs-2006-11259,DBLP:conf/flairs/PuzisGS06,DBLP:conf/iclr/LampleC20,DBLP:journals/corr/abs-2009-03393}. These usually proceed in a backwards fashion by generating random proof trees from a given set of axioms and rules. Wang \emph{et al.} proposed to use supervised learning to learn how to guide this process towards generating more interesting theorems that are similar to those of a reference dataset~\cite{DBLP:conf/nips/WangD20}.

\myparagraph{Using reinforcement learning for loop invariant synthesis.} Reinforcement learning has already been applied to the problem of loop invariant synthesis~\cite{DBLP:conf/nips/SiDRNS18,DBLP:conf/cav/SiNDNS20}, albeit in a very different way. In the aforementioned work (which introduces the Code2Inv benchmark), a \emph{separate} reinforcement learning agent is trained from scratch on every benchmark problem, using counterexamples from an SMT solver as a training signal. This process takes minutes to hours for each problem to be solved. In contrast, we train a \emph{single} agent to generalize across problem instances so that new problems can be solved in a matter of milliseconds.

\myparagraph{Using reinforcement learning for theorem proving.} The HOList Zero~\cite{DBLP:journals/corr/abs-1905-10501} and TacticZero~\cite{DBLP:conf/nips/WuNWD21} systems use reinforcement learning to learn how to interact with tactic-based theorem provers without relying on human proofs. However, they still rely on large human-produced corpora of formalized mathematical statements to be used as training tasks and are not yet competitive with approaches based on imitation learning. The use of reinforcement learning has also been explored in the context of saturation-based or tableau-based provers for first-order logic~\cite{DBLP:conf/nips/KaliszykUMO18,DBLP:conf/aaai/CrouseAMWCKSTWF21}. However, learned heuristics in such settings must operate at a very low-level and are subject to a speed-accuracy tradeoff that is unfavorable to deep learning.


\section{Conclusions and Future Work}

This paper took a bold stride toward the challenge of self-learning automated theorem proving without relying on example theorems and proofs. It advocates a hybrid approach to theorem proving where experts are provided a flexible language to formalize their domain-specific knowledge in the form of generic nondeterministic teacher and solver strategies, leaving blanks to be filled by learning. 

We demonstrated our framework by applying it to the problem of loop invariant synthesis. Our agent solves all Code2Inv challenges. More importantly, it learns to solve a majority of problems with no search at all despite never seeing these problems during training.

None of our core contributions, however, are specific to invariant synthesis. These include:
\begin{inparaenum}[\it i)]
\item our key insight that \emph{nondeterministic programming} and \emph{reinforcement learning} can be similarly combined to implement solvers and teachers,
\item \emph{conditional generative strategies} as a general template to write teachers,
\item \emph{abductive reasoning} as a design principle that is made scalable by self-learned guidance,
\item \emph{strategy events} for easier reward engineering and better sample-efficiency, and
\item an implementation that establishes engineering foundations for making the whole approach practical.
\end{inparaenum}

We see this paper as a first step toward a broader vision where interactive theorem provers allow users to write teacher and solver strategies for domains of mathematics and programming in a modular and distributed fashion. A unique neural network can then be trained as an oracle for all these strategies, allowing better generalization and sample efficiency. However, many challenges lie ahead. In particular, while there is some evidence already~\cite{selsambeyond,gowersmanifesto,DBLP:conf/frocos/EchenimPS19} that effective solver strategies can be written at scale (interactive theorem provers already allow specifying custom search tactics~\cite{DBLP:conf/lpar/Delahaye00,DBLP:books/daglib/0035083,DBLP:conf/itp/FultonMBP17} and we are merely suggesting a generalization of this popular feature), scaling up teacher strategies raises more questions. For example, writing conditional generative strategies can be engineering-intensive since diversity is enforced extrinsically via manually-defined constraints. Future work will explore teachers that also integrate intrinsic reward signals, either via a curiosity mechanism or by having the solver directly reward the teacher for producing problems of suitable novelty and difficulty.


\section*{Broader Impact}

Automated theorem proving for program verification and program synthesis plays an important role for increasing the safety of critical pieces of software. More automatic theorem provers enable a wider user base to benefit from the advances to their software quality.
But automating theorem provers is a highly sophisticated challenge.
Learning from succinct expert advice in the form of strategies may be the sweet spot enabling more generic domain-specific automation for theorem provers.



\section*{Acknowledgments and Disclosure of Funding}

This work was supported by the Alexander von Humboldt Professorship program, the National Science Foundation under Award CNS-1739629 and by DARPA under Contract No.\ FA8750-18-C-0092.
We thank Ruben Martins, Paul Gölz, Vincent Hellendoorn, Léonard Blier along with the anonymous reviewers at NeurIPS 2022 for their helpful comments and feedback.


\bibliographystyle{unsrt}  
\bibliography{main}

\begin{thebibliography}{10}

\bibitem{DBLP:conf/iclr/HuangDSS19}
Daniel Huang, Prafulla Dhariwal, Dawn Song, and Ilya Sutskever.
\newblock {GamePad:} {A} learning environment for theorem proving.
\newblock In {\em 7th International Conference on Learning Representations,
  {ICLR} 2019, New Orleans, LA, USA, May 6-9, 2019}. OpenReview.net, 2019.

\bibitem{DBLP:conf/icml/YangD19}
Kaiyu Yang and Jia Deng.
\newblock Learning to prove theorems via interacting with proof assistants.
\newblock In Kamalika Chaudhuri and Ruslan Salakhutdinov, editors, {\em
  Proceedings of the 36th International Conference on Machine Learning, {ICML}
  2019, 9-15 June 2019, Long Beach, California, {USA}}, volume~97 of {\em
  Proceedings of Machine Learning Research}, pages 6984--6994. {PMLR}, 2019.

\bibitem{DBLP:conf/icml/BansalLRSW19}
Kshitij Bansal, Sarah~M. Loos, Markus~N. Rabe, Christian Szegedy, and Stewart
  Wilcox.
\newblock {HOList:} an environment for machine learning of higher order logic
  theorem proving.
\newblock In Kamalika Chaudhuri and Ruslan Salakhutdinov, editors, {\em
  Proceedings of the 36th International Conference on Machine Learning, {ICML}
  2019, 9-15 June 2019, Long Beach, California, {USA}}, volume~97 of {\em
  Proceedings of Machine Learning Research}, pages 454--463. {PMLR}, 2019.

\bibitem{DBLP:conf/iclr/LiYWP21}
Wenda Li, Lei Yu, Yuhuai Wu, and Lawrence~C. Paulson.
\newblock {IsarStep:} a benchmark for high-level mathematical reasoning.
\newblock In {\em 9th International Conference on Learning Representations,
  {ICLR} 2021, Virtual Event, Austria, May 3-7, 2021}. OpenReview.net, 2021.

\bibitem{DBLP:conf/iclr/HanRWAP22}
Jesse~Michael Han, Jason Rute, Yuhuai Wu, Edward~W. Ayers, and Stanislas Polu.
\newblock Proof artifact co-training for theorem proving with language models.
\newblock In {\em The Tenth International Conference on Learning
  Representations, {ICLR} 2022, Virtual Event, April 25-29, 2022}.
  OpenReview.net, 2022.

\bibitem{DBLP:conf/nips/WangD20}
Mingzhe Wang and Jia Deng.
\newblock Learning to prove theorems by learning to generate theorems.
\newblock In Hugo Larochelle, Marc'Aurelio Ranzato, Raia Hadsell,
  Maria{-}Florina Balcan, and Hsuan{-}Tien Lin, editors, {\em Advances in
  Neural Information Processing Systems 33: Annual Conference on Neural
  Information Processing Systems 2020, NeurIPS 2020, December 6-12, 2020,
  virtual}, 2020.

\bibitem{silver2018general}
David Silver, Thomas Hubert, Julian Schrittwieser, Ioannis Antonoglou, Matthew
  Lai, Arthur Guez, Marc Lanctot, Laurent Sifre, Dharshan Kumaran, Thore
  Graepel, et~al.
\newblock A general reinforcement learning algorithm that masters chess, shogi,
  and go through self-play.
\newblock {\em Science}, 362(6419):1140--1144, 2018.

\bibitem{DBLP:journals/corr/abs-1905-10501}
Kshitij Bansal, Sarah~M. Loos, Markus~N. Rabe, and Christian Szegedy.
\newblock Learning to reason in large theories without imitation.
\newblock {\em CoRR}, abs/1905.10501, 2019.

\bibitem{DBLP:conf/nips/WuNWD21}
Minchao Wu, Michael Norrish, Christian Walder, and Amir Dezfouli.
\newblock {TacticZero:} learning to prove theorems from scratch with deep
  reinforcement learning.
\newblock In Marc'Aurelio Ranzato, Alina Beygelzimer, Yann~N. Dauphin, Percy
  Liang, and Jennifer~Wortman Vaughan, editors, {\em Advances in Neural
  Information Processing Systems 34: Annual Conference on Neural Information
  Processing Systems 2021, NeurIPS 2021, December 6-14, 2021, virtual}, pages
  9330--9342, 2021.

\bibitem{selsambeyond}
Daniel Selsam.
\newblock Beyond the tactic-state automaton.
\newblock {\em Mathematical Reasoning in General Artificial Intelligence
  Workshop, ICLR}, 2021.

\bibitem{DBLP:journals/jar/HahnleS94}
Reiner H\"ahnle and Peter~H. Schmitt.
\newblock The liberalized $\delta$-rule in free variable semantic tableaux.
\newblock {\em J. Autom. Reasoning}, 13(2):211--221, 1994.

\bibitem{DBLP:journals/siamcomp/Cook78}
Stephen~A. Cook.
\newblock Soundness and completeness of an axiom system for program
  verification.
\newblock {\em SIAM J. Comput.}, 7(1):70--90, 1978.

\bibitem{DBLP:books/sp/Harel79}
David Harel.
\newblock {\em First-Order Dynamic Logic}, volume~68 of {\em LNCS}.
\newblock Springer, 1979.

\bibitem{DBLP:conf/cade/Filliatre13}
Jean{-}Christophe Filli{\^{a}}tre.
\newblock One logic to use them all.
\newblock In Maria~Paola Bonacina, editor, {\em Automated Deduction - {CADE-24}
  - 24th International Conference on Automated Deduction, Lake Placid, NY, USA,
  June 9-14, 2013. Proceedings}, volume 7898 of {\em LNCS}, pages 1--20.
  Springer, 2013.

\bibitem{DBLP:journals/cacm/Hoare69}
C.~A.~R. Hoare.
\newblock An axiomatic basis for computer programming.
\newblock {\em Commun. {ACM}}, 12(10):576--580, 1969.

\bibitem{meyer2010abduction}
Michael Meyer.
\newblock Abduction -- a logical view for investigating and initiating
  processes of discovering mathematical coherences.
\newblock {\em Educational Studies in Mathematics}, 74(2):185--205, 2010.

\bibitem{polya2004solve}
George Polya.
\newblock {\em How to solve it: A new aspect of mathematical method},
  volume~85.
\newblock Princeton university press, 2004.

\bibitem{DBLP:conf/oopsla/DilligDLM13}
Isil Dillig, Thomas Dillig, Boyang Li, and Kenneth~L. McMillan.
\newblock Inductive invariant generation via abductive inference.
\newblock In Antony~L. Hosking, Patrick~Th. Eugster, and Cristina~V. Lopes,
  editors, {\em Proceedings of the 2013 {ACM} {SIGPLAN} International
  Conference on Object Oriented Programming Systems Languages {\&}
  Applications, {OOPSLA} 2013, part of {SPLASH} 2013, Indianapolis, IN, USA,
  October 26-31, 2013}, pages 443--456. {ACM}, 2013.

\bibitem{DBLP:conf/frocos/EchenimPS19}
Mnacho Echenim, Nicolas Peltier, and Yanis Sellami.
\newblock {Ilinva:} using abduction to generate loop invariants.
\newblock In Andreas Herzig and Andrei Popescu, editors, {\em Frontiers of
  Combining Systems - 12th International Symposium, FroCoS 2019, London, UK,
  September 4-6, 2019, Proceedings}, volume 11715 of {\em LNCS}, pages 77--93.
  Springer, 2019.

\bibitem{DBLP:journals/jacm/EiterG95}
Thomas Eiter and Georg Gottlob.
\newblock The complexity of logic-based abduction.
\newblock {\em J. {ACM}}, 42(1):3--42, 1995.

\bibitem{DBLP:conf/iclr/DanihelkaGSS22}
Ivo Danihelka, Arthur Guez, Julian Schrittwieser, and David Silver.
\newblock Policy improvement by planning with {Gumbel}.
\newblock In {\em The Tenth International Conference on Learning
  Representations, {ICLR} 2022, Virtual Event, April 25-29, 2022}.
  OpenReview.net, 2022.

\bibitem{DBLP:journals/tciaig/BrownePWLCRTPSC12}
Cameron Browne, Edward~Jack Powley, Daniel Whitehouse, Simon~M. Lucas, Peter~I.
  Cowling, Philipp Rohlfshagen, Stephen Tavener, Diego~Perez Liebana, Spyridon
  Samothrakis, and Simon Colton.
\newblock A survey of {Monte Carlo Tree Search} methods.
\newblock {\em {IEEE} Trans. Comput. Intell. {AI} Games}, 4(1):1--43, 2012.

\bibitem{DBLP:conf/iclr/BunelHDSK18}
Rudy Bunel, Matthew~J. Hausknecht, Jacob Devlin, Rishabh Singh, and Pushmeet
  Kohli.
\newblock Leveraging grammar and reinforcement learning for neural program
  synthesis.
\newblock In {\em 6th International Conference on Learning Representations,
  {ICLR} 2018, Vancouver, BC, Canada, April 30 - May 3, 2018, Conference Track
  Proceedings}. OpenReview.net, 2018.

\bibitem{DBLP:conf/iclr/LampleC20}
Guillaume Lample and Fran{\c{c}}ois Charton.
\newblock Deep learning for symbolic mathematics.
\newblock In {\em 8th International Conference on Learning Representations,
  {ICLR} 2020, Addis Ababa, Ethiopia, April 26-30, 2020}. OpenReview.net, 2020.

\bibitem{DBLP:journals/corr/abs-1902-10565}
David~J. Wu.
\newblock Accelerating self-play learning in go.
\newblock {\em CoRR}, abs/1902.10565, 2019.

\bibitem{DBLP:conf/nips/PaszkeGMLBCKLGA19}
Adam Paszke, Sam Gross, Francisco Massa, Adam Lerer, James Bradbury, Gregory
  Chanan, Trevor Killeen, Zeming Lin, Natalia Gimelshein, Luca Antiga, Alban
  Desmaison, Andreas K{\"{o}}pf, Edward~Z. Yang, Zachary DeVito, Martin Raison,
  Alykhan Tejani, Sasank Chilamkurthy, Benoit Steiner, Lu~Fang, Junjie Bai, and
  Soumith Chintala.
\newblock Pytorch: An imperative style, high-performance deep learning library.
\newblock In Hanna~M. Wallach, Hugo Larochelle, Alina Beygelzimer, Florence
  d'Alch{\'{e}}{-}Buc, Emily~B. Fox, and Roman Garnett, editors, {\em Advances
  in Neural Information Processing Systems 32: Annual Conference on Neural
  Information Processing Systems 2019, NeurIPS 2019, December 8-14, 2019,
  Vancouver, BC, Canada}, pages 8024--8035, 2019.

\bibitem{DBLP:conf/osdi/MoritzNWTLLEYPJ18}
Philipp Moritz, Robert Nishihara, Stephanie Wang, Alexey Tumanov, Richard Liaw,
  Eric Liang, Melih Elibol, Zongheng Yang, William Paul, Michael~I. Jordan, and
  Ion Stoica.
\newblock Ray: {A} distributed framework for emerging {AI} applications.
\newblock In Andrea~C. Arpaci{-}Dusseau and Geoff Voelker, editors, {\em 13th
  {USENIX} Symposium on Operating Systems Design and Implementation, {OSDI}
  2018, Carlsbad, CA, USA, October 8-10, 2018}, pages 561--577. {USENIX}
  Association, 2018.

\bibitem{DBLP:journals/corr/Hedges14a}
Jules Hedges.
\newblock Monad transformers for backtracking search.
\newblock In Paul~Blain Levy and Neel Krishnaswami, editors, {\em Proceedings
  5th Workshop on Mathematically Structured Functional Programming, MSFP@ETAPS
  2014, Grenoble, France, 12 April 2014}, volume 153 of {\em {EPTCS}}, pages
  31--50, 2014.

\bibitem{DBLP:conf/nips/SiDRNS18}
Xujie Si, Hanjun Dai, Mukund Raghothaman, Mayur Naik, and Le~Song.
\newblock Learning loop invariants for program verification.
\newblock In {\em Advances in Neural Information Processing Systems}, pages
  7762--7773, 2018.

\bibitem{DBLP:conf/cav/SiNDNS20}
Xujie Si, Aaditya Naik, Hanjun Dai, Mayur Naik, and Le~Song.
\newblock Code2inv: {A} deep learning framework for program verification.
\newblock In Shuvendu~K. Lahiri and Chao Wang, editors, {\em Computer Aided
  Verification - 32nd International Conference, {CAV} 2020, Los Angeles, CA,
  USA, July 21-24, 2020, Proceedings, Part {II}}, volume 12225 of {\em LNCS},
  pages 151--164. Springer, 2020.

\bibitem{DBLP:conf/iclr/RyanWYGJ20}
Gabriel Ryan, Justin Wong, Jianan Yao, Ronghui Gu, and Suman Jana.
\newblock {CLN2INV:} learning loop invariants with continuous logic networks.
\newblock In {\em 8th International Conference on Learning Representations,
  {ICLR} 2020, Addis Ababa, Ethiopia, April 26-30, 2020}. OpenReview.net, 2020.

\bibitem{selsam2020universal}
Daniel Selsam, Jesse~Michael Han, Leonardo de~Moura, and Patrice Godefroid.
\newblock Universal policies for software-defined {MDPs}.
\newblock {\em arXiv preprint arXiv:2012.11401}, 2020.

\bibitem{selsamAITPTalk}
Daniel Selsam.
\newblock The {IMO} grand challenge.
\newblock \url{http://aitp-conference.org/2020/slides/DS.pdf}, 2020.
\newblock Talk at AITP 2020.

\bibitem{selsamStrategyML}
Daniel Selsam.
\newblock The {IMO} grand challenge: A battle of ideas.
\newblock \url{https://dselsam.github.io/IMO-GC-battle-of-ideas/}, 2020.

\bibitem{DBLP:conf/iclr/WuJBG21}
Yuhuai Wu, Albert~Q. Jiang, Jimmy Ba, and Roger~Baker Grosse.
\newblock {INT:} an inequality benchmark for evaluating generalization in
  theorem proving.
\newblock In {\em 9th International Conference on Learning Representations,
  {ICLR} 2021, Virtual Event, Austria, May 3-7, 2021}. OpenReview.net, 2021.

\bibitem{DBLP:journals/corr/abs-2006-11259}
Eser Ayg{\"{u}}n, Zafarali Ahmed, Ankit Anand, Vlad Firoiu, Xavier Glorot,
  Laurent Orseau, Doina Precup, and Shibl Mourad.
\newblock Learning to prove from synthetic theorems.
\newblock {\em CoRR}, abs/2006.11259, 2020.

\bibitem{DBLP:conf/flairs/PuzisGS06}
Yury Puzis, Yi~Gao, and Geoff Sutcliffe.
\newblock Automated generation of interesting theorems.
\newblock In Geoff Sutcliffe and Randy Goebel, editors, {\em Proceedings of the
  Nineteenth International Florida Artificial Intelligence Research Society
  Conference, Melbourne Beach, Florida, USA, May 11-13, 2006}, pages 49--54.
  {AAAI} Press, 2006.

\bibitem{DBLP:journals/corr/abs-2009-03393}
Stanislas Polu and Ilya Sutskever.
\newblock Generative language modeling for automated theorem proving.
\newblock {\em CoRR}, abs/2009.03393, 2020.

\bibitem{DBLP:conf/nips/KaliszykUMO18}
Cezary Kaliszyk, Josef Urban, Henryk Michalewski, and Miroslav Ols{\'{a}}k.
\newblock Reinforcement learning of theorem proving.
\newblock In Samy Bengio, Hanna~M. Wallach, Hugo Larochelle, Kristen Grauman,
  Nicol{\`{o}} Cesa{-}Bianchi, and Roman Garnett, editors, {\em Advances in
  Neural Information Processing Systems 31: Annual Conference on Neural
  Information Processing Systems 2018, NeurIPS 2018, December 3-8, 2018,
  Montr{\'{e}}al, Canada}, pages 8836--8847, 2018.

\bibitem{DBLP:conf/aaai/CrouseAMWCKSTWF21}
Maxwell Crouse, Ibrahim Abdelaziz, Bassem Makni, Spencer Whitehead, Cristina
  Cornelio, Pavan Kapanipathi, Kavitha Srinivas, Veronika Thost, Michael
  Witbrock, and Achille Fokoue.
\newblock A deep reinforcement learning approach to first-order logic theorem
  proving.
\newblock In {\em Thirty-Fifth {AAAI} Conference on Artificial Intelligence,
  {AAAI} 2021, Thirty-Third Conference on Innovative Applications of Artificial
  Intelligence, {IAAI} 2021, The Eleventh Symposium on Educational Advances in
  Artificial Intelligence, {EAAI} 2021, Virtual Event, February 2-9, 2021},
  pages 6279--6287. {AAAI} Press, 2021.

\bibitem{gowersmanifesto}
Timothy Gowers.
\newblock How can it be feasible to find proofs?
\newblock
  \url{https://gowers.wordpress.com/2022/04/28/announcing-an-automatic-theorem-proving-project/},
  2022.

\bibitem{DBLP:conf/lpar/Delahaye00}
David Delahaye.
\newblock A tactic language for the system {Coq}.
\newblock In Michel Parigot and Andrei Voronkov, editors, {\em Logic for
  Programming and Automated Reasoning, 7th International Conference, {LPAR}
  2000, Reunion Island, France, November 11-12, 2000, Proceedings}, volume 1955
  of {\em LNCS}, pages 85--95. Springer, 2000.

\bibitem{DBLP:books/daglib/0035083}
Adam Chlipala.
\newblock {\em Certified Programming with Dependent Types - {A} Pragmatic
  Introduction to the Coq Proof Assistant}.
\newblock {MIT} Press, 2013.

\bibitem{DBLP:conf/itp/FultonMBP17}
Nathan Fulton, Stefan Mitsch, Brandon Bohrer, and Andr{\'e} Platzer.
\newblock Bellerophon: Tactical theorem proving for hybrid systems.
\newblock In Mauricio Ayala-Rinc{\'o}n and C{\'e}sar~A. Mu{\~n}oz, editors,
  {\em ITP}, volume 10499 of {\em LNCS}, pages 207--224. Springer, 2017.

\bibitem{DBLP:journals/jct/DantzigE73}
George~B. Dantzig and B.~Curtis Eaves.
\newblock {Fourier-Motzkin} elimination and its dual.
\newblock {\em J. Comb. Theory, Ser. {A}}, 14(3):288--297, 1973.

\bibitem{DBLP:conf/nips/VaswaniSPUJGKP17}
Ashish Vaswani, Noam Shazeer, Niki Parmar, Jakob Uszkoreit, Llion Jones,
  Aidan~N. Gomez, Lukasz Kaiser, and Illia Polosukhin.
\newblock Attention is all you need.
\newblock In Isabelle Guyon, Ulrike von Luxburg, Samy Bengio, Hanna~M. Wallach,
  Rob Fergus, S.~V.~N. Vishwanathan, and Roman Garnett, editors, {\em Advances
  in Neural Information Processing Systems 30: Annual Conference on Neural
  Information Processing Systems 2017, December 4-9, 2017, Long Beach, CA,
  {USA}}, pages 5998--6008, 2017.

\bibitem{Fey/Lenssen/2019}
Matthias Fey and Jan~E. Lenssen.
\newblock Fast graph representation learning with {PyTorch Geometric}.
\newblock In {\em ICLR Workshop on Representation Learning on Graphs and
  Manifolds}, 2019.

\bibitem{DBLP:conf/iclr/HellendoornSSMB20}
Vincent~J. Hellendoorn, Charles Sutton, Rishabh Singh, Petros Maniatis, and
  David Bieber.
\newblock Global relational models of source code.
\newblock In {\em 8th International Conference on Learning Representations,
  {ICLR} 2020, Addis Ababa, Ethiopia, April 26-30, 2020}. OpenReview.net, 2020.

\bibitem{DBLP:conf/nips/ShivQ19}
Vighnesh~Leonardo Shiv and Chris Quirk.
\newblock Novel positional encodings to enable tree-based transformers.
\newblock In Hanna~M. Wallach, Hugo Larochelle, Alina Beygelzimer, Florence
  d'Alch{\'{e}}{-}Buc, Emily~B. Fox, and Roman Garnett, editors, {\em Advances
  in Neural Information Processing Systems 32: Annual Conference on Neural
  Information Processing Systems 2019, NeurIPS 2019, December 8-14, 2019,
  Vancouver, BC, Canada}, pages 12058--12068, 2019.

\bibitem{DBLP:conf/iclr/HahnSKRF21}
Christopher Hahn, Frederik Schmitt, Jens~U. Kreber, Markus~Norman Rabe, and
  Bernd Finkbeiner.
\newblock Teaching temporal logics to neural networks.
\newblock In {\em 9th International Conference on Learning Representations,
  {ICLR} 2021, Virtual Event, Austria, May 3-7, 2021}. OpenReview.net, 2021.

\end{thebibliography}



\newpage
\appendix
\onecolumn
\section{Details on the Solver Strategy}\label{ap:solver-details}

In Figure~\ref{fig:solver-code}, we provide some detailed pseudocode for the solver strategy that we implemented for the purpose of the experiments in Section~\ref{sec:experiments}. This strategy is similar to the simple one introduced in Figure~\ref{fig:simplified-solver} but it comes with the following extensions:

\begin{itemize}
  \item \textbf{Ability to abduct disjunctive invariants:} Imagine a proof obligation fails and the \texttt{abduct} function returns $a_1, \dots, a_n$ as possible missing assumptions. In some cases, none of those can be proved invariant but the disjunction of a subset of them can. This is why the \texttt{suggest\_missing} function defined on line~\ref{line:suggest-missing} is used to build an invariant candidate as a disjunction of at most 3 (atomic) abduction candidates.
  \item \textbf{Ability to strengthen abducted invariants:} An abducted invariant candidate may have to be strengthened before it can be proved invariant. For example, the abduction engine may suggest $x \neq 0$ as a missing assumption but $x \neq 0$ may not be a valid invariant whereas $x > 0$ (or $x < 0$) is. The call to \texttt{strengthen} on line~\ref{line:strengthen} is used to optionally and nondeterministically strengthen an invariant candidate. Our current strategy supports two kinds of strengthening: replacing a formula of the form $A \neq B$ by either $A > B$ or $A < B$ or weakening an inequality of the form $A \leq B$ into $A \leq B + c_{?}$ where $c_{?}$ is a nonnegative constant whose exact value is to be determined later (see next point).
  \item \textbf{Ability to instantiate constants lazily via abduction:} Invariant candidates can feature metavariables that denote unknown constants. The \texttt{constrs} variable defined in line~\ref{line:constrs} collects global constraints about these metavariables. If a missing assumption is suggested that only features metavariables, then it is added as a constraint rather than as a new invariant candidate (line~\ref{line:meta-only}). After an invariant is proved, the metavariables it contains are instantiated using concrete values in a way to satisfy all global constraints (see call to \texttt{abduct\_refinement} on line~\ref{line:abduct-refinement}). A metavariable that appears in the invariant as an upper bound is instantiated with a value that is as low as possible, whereas a metavariable that appears as a lower bound is instantiated with a value that is as high as possible (this information is contained in the \texttt{btype} variable).
  \item \textbf{Ability to conjecture invariant templates:}
    In some cases, abduction alone is not enough to discover a missing invariant. For example, if two disjunctive invariants $I_1$ and $I_2$ are needed for proving the postcondition, one may have to conjecture the first one and then use abduction to find the other. The strategy in Figure~\ref{fig:solver-code} allows conjecturing invariant templates with metavariables to be instantiated via abduction (see previous point). The \texttt{conjectures} function called on line~\ref{line:conjecture} returns three kinds of conjecture candidates:
    \begin{inparaenum}[\it i)]
      \item conjectures of the form $t \odot c_?$ where $\odot \in \{\ge, =\}$ and $t = \sum_i a_ix_i$ is a linear combination of variables that is preserved by the loop body,
      \item relaxations of the loop guard (e.g. $x \le 10 + c_?$ with $c_? > 0$ if the loop guard is $x \le 10$) and
      \item initial assumptions that only contain variables that are not modified by the loop body.
    \end{inparaenum}
\end{itemize}

The solver strategy also emits two types of events (see Section~\ref{sec:vec-rewards}) at lines~\ref{line:abduction-ev} and \ref{line:conjecturing-ev} respectively. Conjecturing events are associated with a reward of $-0.3$ and abduction events are associated with a reward of $-0.2$. Both kinds of events are counted at most four times ($m_e = 4$) and the minimum total reward delivered in case of a success is $\rmin = 0$.

Hints on how to use our proposed solver strategy to solve Code2Inv benchmark problems are available in Appendix~\ref{ap:code2inv-examples}.

\begin{figure}
  \setlstpy{}
  \input{content/solver-strategy.tex}
  \caption{Strategy for the solver agent. The code in gray is only useful for providing the network with contextual information and can be disregarded on first reading. Every call to \texttt{choose} is implicitly passed the program counter along with the value of all global parameters and all variables that are defined in lines~\ref{line:net-start} to \ref{line:net-end}. In particular, the \texttt{pending} variable summarizes all information from the program stack that is relevant to the neural network. }\label{fig:solver-code}
\end{figure}


\section{Details on the Teacher Strategy}\label{ap:teacher-details}

The teacher strategy we use for loop invariant generation follows the structure introduced in Figure~\ref{fig:teacher-simple}. We provide additional details below:

\begin{itemize}
\item \textbf{Sampling constraints:} The full list of available constraints is available in Table~\ref{tab:teacher-constrs-full}. Values are sampled \emph{mostly} independently for each constraint types. In our implementation, we hardcode a small number of correlations (e.g. we are more likely to sample a disjunctive formula for \texttt{inv\_main} if \texttt{inv\_lin} is not used in order to keep things interesting). We also reject a number of constraint combinations that are clearly uninteresting or unsatisfiable.
\item \textbf{Fixed constraints:} In addition to penalizing the violation of sampled constraints, the teacher implements fixed hard constraints that are always enforced. A list of all such constraints is available in Table~\ref{tab:teacher-fixed-constraints}. Violating one of these constraints leads to an immediate failure along with a reward of -1.
\item \textbf{Refining formulas and using abduction:} Different parts of the problem template shown in Figure~\ref{fig:teacher-simple} are nondeterministically refined in turn. To refine an atomic formula, a template is first selected of the form $x_? \odot c_?$ or $x_? \odot y_?$ where $x_?$ and $y_?$ are variable placeholders, $c_?$ is a constant placeholder and $\odot \in \{<, \le, >, \ge, =, \neq\}$. Each variable placeholder is then nondeterministically substituted by an existing or a fresh variable. Constant placeholders are either instantiated with concrete constants (a set of 6 available numerical constants is sampled at the start of the teacher strategy along with constraints) or parameters (special variables that cannot be modified by the program). Constant placeholders can also be left as-is and refined later using abduction (e.g. before invariant preservation is checked, abduction is used to suggest values for the remaining constant placeholders). Abduction is also used to suggest required parameter assumptions that are added to \texttt{init} (e.g. $n > 0$ where $n$ is a variable not modified in the program).
\item \textbf{Refining programs:} Subprograms are refined by selecting a sequence of assignment templates of the form: $x_? := c_?$, $x_? := y_?$, $x_? := x_? + d_?$, $x_? := x_? - d_?$, $x_? := x_? + y_?$ and $x_? := c_? - y_?$ ($d_?$ is a placeholder for a strictly positive constant). A special \texttt{skip} template can be selected to stop adding assignments. Variable and constant placeholders are handled in the same way they are handled in formulas. Which templates are available is determined by the \texttt{assignment-templates} constraint (see Figure~\ref{tab:teacher-constrs-full}).
\item \textbf{Extra refinement suggestions:} Extra suggestions are added to the standard templates when refining some formulas. When adding a disjunct to the main invariant, the loop guard itself is added as a suggestion along with a relaxed version of it (using a placeholder constant). When refining the last disjunct of the postcondition, abduction is used to suggest candidates that are consequences of the current assumptions in the associated proof obligation. Abduction is also used to suggest conjuncts for \texttt{init}.
\item \textbf{Detecting constraint violations early:} Constraint violations are detected as early as possible to allow early feedback during search. For example, whether or not the invariant is satisfiable is checked right after the invariant has been refined and before the loop body is refined in turn. Most constraints must be checked again whenever a new parameter assumption is added. For example, an invariant $0 \le x \wedge x < n$ may be initially judged as satisfiable. However, abduction may later on suggest $n < 0$ as a parameter assumption, making it unsatisfiable.
\item \textbf{Applying random transformations to generated problems:} For increased diversity, a sequence of random transformations is applied to any problem generated by the teacher before it is returned. We provide a list of all such transformations in Table~\ref{tab:teacher-transformations}.
\end{itemize}

\begin{table}
  \renewcommand{\arraystretch}{1.5}
  \centering
\small
\newcommand{\boolvals}{\texttt{bool}}
\newcommand{\litvals}[1]{\texttt{#1}}
\newcommand{\tsep}{\hline}
\renewcommand\tabularxcolumn[1]{m{#1}} 
\begin{tabularx}{\textwidth}{m{4.3cm}m{1.5cm}X}
\textbf{Name} & \textbf{Type} & \textbf{Description} \\
\toprule
\texttt{num-preserved-term-vars} & \litvals{none|2|3} &
If an integer $n$, then \texttt{inv\_lin} is refined with an
invariant of the form $\sum_{i=1}^{n} a_ix_i = c$.
\\ \tsep
\texttt{num-inv-main-disjuncts} & \litvals{none|1|2} &
If an integer $n$, then \texttt{inv\_main} is refined with a disjunction of $n$ atomic formulas.
\\ \tsep
\texttt{num-inv-aux-conjuncts} & \litvals{none|1|2} &
If an integer $n$, then \texttt{inv\_aux} is refined with a conjunction of $n$ atomic formulas.
\\ \tsep
\texttt{num-post-disjuncts} & \litvals{1|2} &
Number of desired atomic disjuncts for \texttt{post}.
\\ \tsep
\texttt{has-conditional} & \boolvals{} &
Whether \texttt{body} must include a conditional statement.
\\ \tsep
\texttt{has-else-branch} & \boolvals{} &
Whether the conditional in \texttt{body} has an else branch.
\\ \tsep
\texttt{has-cond-guard} & \boolvals{} &
Whether the conditional in \texttt{body} has a guard. If not, the guard is refined with the nondeterministic expression \texttt{*}.
\\ \tsep
\texttt{body-implies-main-inv} & \boolvals{} &
If \texttt{true}, then \texttt{inv\_main} always holds after executing \texttt{body} regardless of whether or not it holds before.
\\ \tsep
\texttt{loop-guard-useful-for-inv} & \boolvals{} &
Whether assuming the loop guard is useful in proving that \texttt{inv\_main} is preserved.
\\ \tsep
\texttt{loop-guard-useful-for-post} & \boolvals{} &
Whether assuming the negation of the loop guard is useful in proving the postcondition \texttt{post}.
\\ \tsep
\texttt{use-params} & \boolvals{} &
Whether or not to use variables that have a constant value throughout the program.
\\ \tsep
\texttt{eq-only-for-init} & \boolvals{} &
Whether or not to use equalities only in \texttt{init}.
\\ \tsep
\texttt{loop-guard-template} & \texttt{template} &
The template to be used to refine the loop guard (e.g. a constant upper bound on a variable).
\\ \tsep
\texttt{assignment-templates} & \texttt{templates} &
Allowed assignment templates for the body (e.g. constant var increment, assigning a var to another one\dots)
\\ \tsep
\texttt{allow-vcomp-in-inv-main} & \boolvals{} &
Whether \texttt{inv\_main}'s first disjunct can feature a
comparison between two variables modified by the program.
\\ \bottomrule
\end{tabularx}

  \bigskip
  \caption{Complete list of teacher constraints. Every constraint type is associated with a separate violation event. The associated reward is $-0.5$ for all constraint types except the last four ones where it is $-0.2$. A total reward of at least $\rmin=-0.5$ is delivered in case of a success.}\label{tab:teacher-constrs-full}
\end{table}

\begin{table}
  \renewcommand{\arraystretch}{1.5}
  \centering
  \small
\newcommand{\tsep}{\hline}
\renewcommand\tabularxcolumn[1]{m{#1}} 
\begin{tabularx}{\textwidth}{m{4.3cm}X}
\textbf{Name} & \textbf{Description} \\
\hline\tsep
\texttt{correctness} & The problem is correct, meaning that the invariant (i.e. the conjunction of \texttt{inv\_lin}, \texttt{inv\_main} and \texttt{inv\_aux}) respects the three properties defining a valid invariant (i.e holds initially, preserved by the loop body and implies the postcondition). \\ \tsep
\texttt{\{inv\_main,post,init\}-not-valid-unsat-or-redundant} & Disjunctive or conjuctive formulas such as \texttt{inv\_main}, \texttt{post} and \texttt{init} must not be valid, unsatisfiable or redundant in the sense that they can be simplified (e.g. $x > 0 \vee x = 0$ is redundant because it simplifies to $c \ge 0$ and $x > 1 \wedge x > 2$ is redundant because it simplifies to $x > 2$). \\ \tsep
\texttt{inv-sat} & The invariant must be satisfiable. \\ \tsep
\texttt{loop-terminates} & The loop guard must not be preserved by the loop body when assuming the invariant, in which case the loop would never terminate once entered. (Note this constraint only rules out a subset of nontermination cases.) \\ \tsep
\texttt{loop-entered} & The \texttt{init} formula does not imply the negation of the loop guard.  \\ \bottomrule
\end{tabularx}

  \bigskip
  \caption{Table of fixed teacher hard constraints. Violation of such a constraint leads to an immediate failure and to a reward of -1.}\label{tab:teacher-fixed-constraints}
\end{table}

\begin{table}
  \renewcommand{\arraystretch}{1.5}
  \centering
  \small
\newcommand{\tsep}{\hline}
\renewcommand\tabularxcolumn[1]{m{#1}} 
\begin{tabularx}{\textwidth}{m{4.3cm}X}
\textbf{Name} & \textbf{Description} \\ \toprule
\texttt{add-useless-loop-guard} & If the loop guard is irrelevant, assign a random formula in its place. \\ \tsep
\texttt{add-useless-init} & Add a random conjunct to \texttt{init}. \\ \tsep
\texttt{add-useless-post} & Add a random disjunct to \texttt{post}. \\ \tsep
\texttt{add-useless-cond} & Replace \texttt{body} by \texttt{if(cond)\{body\}} where \texttt{cond} is a random formula. \\ \tsep
\texttt{rearrange-commutative} & Shuffle the order of disjunctions and conjunctions. \\ \tsep
\texttt{move-conditional} & Commute the conditional statement in \texttt{body} with other instructions. \\ \tsep
\texttt{shuffle-instrs} & Shuffle the order of consecutive assignments. \\ \tsep
\texttt{randomize-comparisons} & Randomize comparisons by changing variable order or converting strict inequalities to nonstrict inequalities and vice versa. \\ \tsep
\texttt{move-param-assum} & Remove a parameter assumption and add its negation to \texttt{post}. \\ \tsep
\texttt{make-post-assums} & Rewrite a disjunctive final assertion into a sequence of atomic assumptions with a final atomic assertion (e.g. rewrite ``\texttt{assert x>0||y>0}'' into ``\texttt{assume x<=0; assert y>0}''). \\ \tsep
\texttt{make-init-instrs} & Replace the \texttt{init} conjunctive assumption by a sequence of variable assignments ad atomic assumptions. \\ \tsep
\texttt{weaken-post} & Weaken the final (atomic) postcondition (e.g. replace ``\texttt{assert x>0}'' by ``\texttt{assert x!=0}''). \\ \bottomrule
\end{tabularx}

  \bigskip
  \caption{Complete list of the final problem transformations implemented by the teacher. Some transformations may trigger or not based on a fixed probability. A transformation application is cancelled if it leads to violating a hard constraint or increasing the number of soft constraint violations.}\label{tab:teacher-transformations}
\end{table}


\section{Examples of Code2Inv Problems}\label{ap:code2inv-examples}

We show examples of Code2Inv benchmark problems in Table~\ref{tab:code2inv-examples}, along with some hints on how they can be solved using the strategy detailed in Appendix~\ref{ap:solver-details}.

\begin{table}
\setlstc{}
\begin{tabular}{ll}
\hline
\noindent\begin{minipage}{0.25\textwidth}
\begin{lstlisting}
x = 0;
while (x < 5) {
  x = x + 1;
  if (y > z) {
    y = z;
  }
}
assert y <= z;
\end{lstlisting}
\end{minipage}
&
\begin{minipage}{0.7\textwidth}
\textbf{Problem 3} \\[0.5em]
Invariant: $x < 5 \,\vee\, y \le z$.  \\[0.5em]
This problem can be solved using our proposed strategy by directly abducting the correct disjunctive invariant when attempting to prove the postcondition.
\end{minipage} \\
\hline
\noindent\begin{minipage}{0.25\textwidth}
\begin{lstlisting}
assume x <= 10;
assume y >= 0;
while (*) {
  x = x + 10;
  y = y + 10;
}
assume x == 20;
assert y != 0;
\end{lstlisting}
\end{minipage}
&
\begin{minipage}{0.7\textwidth}
\textbf{Problem 7} \\[0.5em]
Invariant: $x - y \le 10$.  \\[0.5em]
The star (\texttt{*}) corresponds to a nondeterministic boolean value. In this case, the loop body can be executed an arbitrary number of times.  \\[0.5em]
This problem can be solved by conjecturing an invariant of the form $x - y \le c_?$ and then using abduction to refine $c_?$.
\end{minipage} \\
\hline
\noindent\begin{minipage}{0.28\textwidth}
\begin{lstlisting}
assume n >= 0;
i = 0;
x = 0;
y = 0;
while (i < n) {
  i = i + 1;
  if (*) {
    x = x + 1;
    y = y + 2;
  } else {
    x = x + 2;
    y = y + 1;
  }
}
assert 3*n == x + y;
\end{lstlisting}
\end{minipage}
&
\begin{minipage}{0.7\textwidth}
\textbf{Problem 93} \\[0.5em]
Invariant: $3i = x + y \,\wedge\, i \le n$. \\[0.5em]
This problem can be solved by conjecturing an invariant of the form $3i-x-y = c_?$, refining $c_?$ through abduction and then abducting $i \le n$ as a missing invariant while trying to prove the postcondition.
\end{minipage} \\
\hline
\noindent\begin{minipage}{0.28\textwidth}
\begin{lstlisting}
s = 0;
i = 1;
while i <= n:
  i = i + 1
  s = s + 1
assume s != 0
assert s == n
\end{lstlisting}
\end{minipage}
&
\begin{minipage}{0.7\textwidth}
\textbf{Problem 110} \\[0.5em]
Invariant: $i-s = 1 \,\wedge\, (i \le n+1 \,\vee\, s = 0)$. \\[0.5em]
This problem can be solved by first conjecturing an invariant of the form $i - s = c_?$, then using abduction to refine $c_?$ and finally abducting the disjunctive invariant $i \le n+1 \,\vee\, s = 0$ while trying to prove the postcondition.
\end{minipage} \\
\hline
\end{tabular}

\bigskip
\caption{Some examples of Code2Inv problems.}\label{tab:code2inv-examples}
\end{table}


\section{Examples of Generated Teacher Problems}\label{ap:teacher-examples}

We show examples of challenge problems generated by our trained teacher agent in Table~\ref{tab:teacher-examples}.

\begin{table}
\setlstc{}
\begin{tabular}{ll}
\hline
\noindent\begin{minipage}{0.5\textwidth}
\begin{lstlisting}
main-inv 1
body-structure no-cond
loop-guard-useful-for-post
available-consts -9 -6 4 8

assume y == x;
while (x < 1) {
    invariant y == x;
    y = y + 1;
    x = x + 1;
}
assert y > 0;
\end{lstlisting}
\end{minipage}
&
\begin{minipage}{0.45\textwidth}
In this example, the network has been tasked to generate an example of a program with a single atomic invariant and a loop guard that is useful to establish the postcondition but not the invariant itself.
\end{minipage} \\
\hline
\noindent\begin{minipage}{0.5\textwidth}
\begin{lstlisting}
main-inv 1
use-aux-inv 2
body-structure no-cond
allow-vcomp-in-prim-inv
no-var-const-assign
available-consts -8 -7 2 6

assume x < y;
assume x >= 5;
while (*) {
    invariant x < y;
    invariant y >= 6 && x >= -1;
    x = x + 6;
    y = y + x;
}
assert x < y;
\end{lstlisting}
\end{minipage}
&
\begin{minipage}{0.45\textwidth}
In this example, the network has been tasked to generate an example that involves an auxilliary event with two conjuncts. The auxilliary event can be useful to prove the main invariant but not the postcondition. Assignments of the form $x = y$ or $x = c$ where $x$ and $y$ are variables and $c$ is a constant are not allowed.
\end{minipage} \\
\hline
\noindent\begin{minipage}{0.5\textwidth}
\begin{lstlisting}
preserved-term 2
disjunctive-post
body-structure no-cond
only-constr-incr
available-consts -55 -52 21 21

assume x == 21;
assume y == -52;
while (*) {
    invariant -x - 3*y == 135;
    y = y - 21;
    x = x + 63;
}
assert y != 21 || x == -198;
\end{lstlisting}
\end{minipage}
&
\begin{minipage}{0.45\textwidth}
In this example, the network has been tasked to find an example of a problem involving a linear invariant with two variables, no loop guard and a disjunctive postcondition. Note that an irrelevant loop guard may be added in the final transformation stage of the teacher.
\end{minipage} \\
\hline
\end{tabular}

\bigskip
\caption{Some examples of problems generated by the teacher. We show the associated invariants for clarity but those should of course be hidden before problems are sent to the solver agent. These invariants only provide one way to solve the associated problems and alternative invariants may exist. We disable the final random transformations applied by the teacher for clarity and to avoid clutter from useless formulas and instructions. Finally, some problem constraints are not listed for brevity unless their value is different from the default with highest probability.}\label{tab:teacher-examples}
\end{table}


\section{\looprl{} UI Screenshots}\label{ap:screenshots}

\begin{figure}
\begin{subfigure}{\textwidth}
\includegraphics[width=\textwidth]{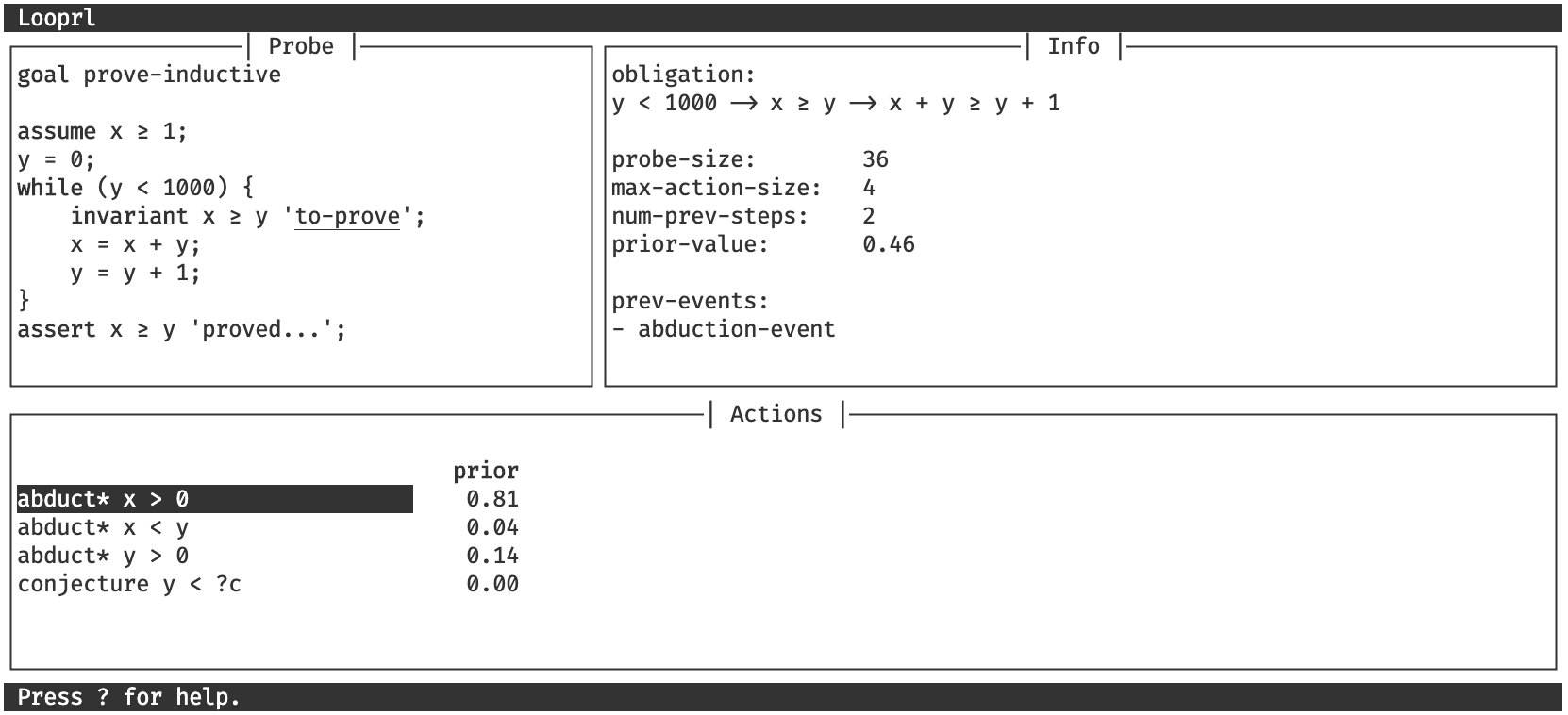}
\caption{Showing the proof obligation associated with the abduction call along with the neural network policy prior.}\label{subfig:solver-obligation}
\end{subfigure}
\begin{subfigure}{\textwidth}
\includegraphics[width=\textwidth]{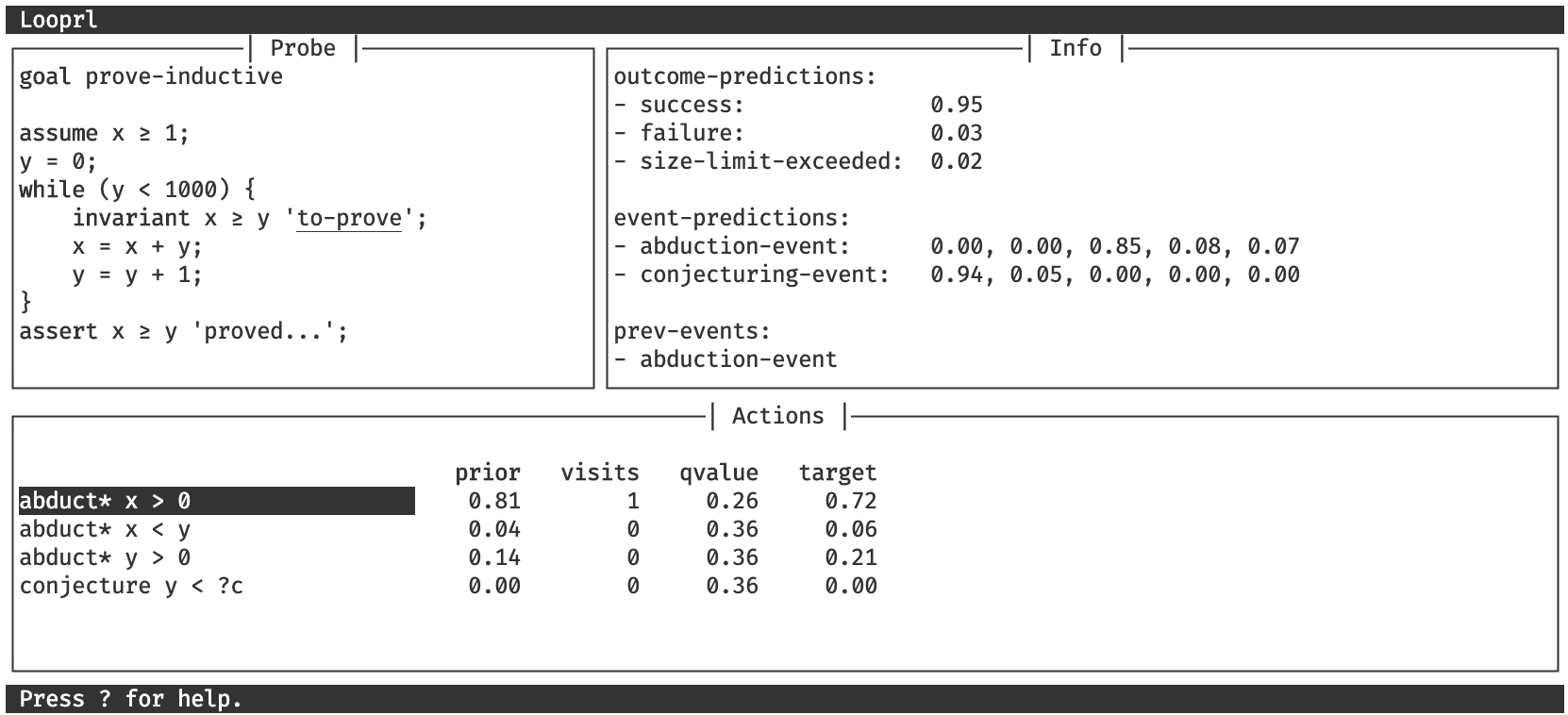}
\caption{Showing event predictions along with MCTS statistics.}\label{subfig:solver-events}
\end{subfigure}
\caption{Visualizing the solver strategy with the \looprl{} UI. In the screenshots above, the UI is used to examine a choice point in the solver strategy where the current invariant candidate $x \ge y$ cannot be proved to be inductive and the user must choose between proving one of several abduction candidates or making a conjecture (this roughly corresponds to the call to \texttt{choose} on line~\ref{line:choose-suggs}). Here, one can see the network assigning a high prior probability to the optimal proof action, which is to try and prove $x > 0$ as an invariant. The network also predicts a value of $0.46$ for this state, which is close to the truth of $0.4$ (proving this problem requires three abduction events with cost $0.2$ each). The details of how the value is estimated can be consulted in screenshot~\ref{subfig:solver-events}. Here, we can see that the network predicts a $0.95$ probability of success along with a probability of $0.94$ for not requiring any conjecture and a probability of $0.85$ for needing two more abduction events.}\label{fig:solver-ap-ui}
\end{figure}

\begin{figure}
\includegraphics[width=\textwidth]{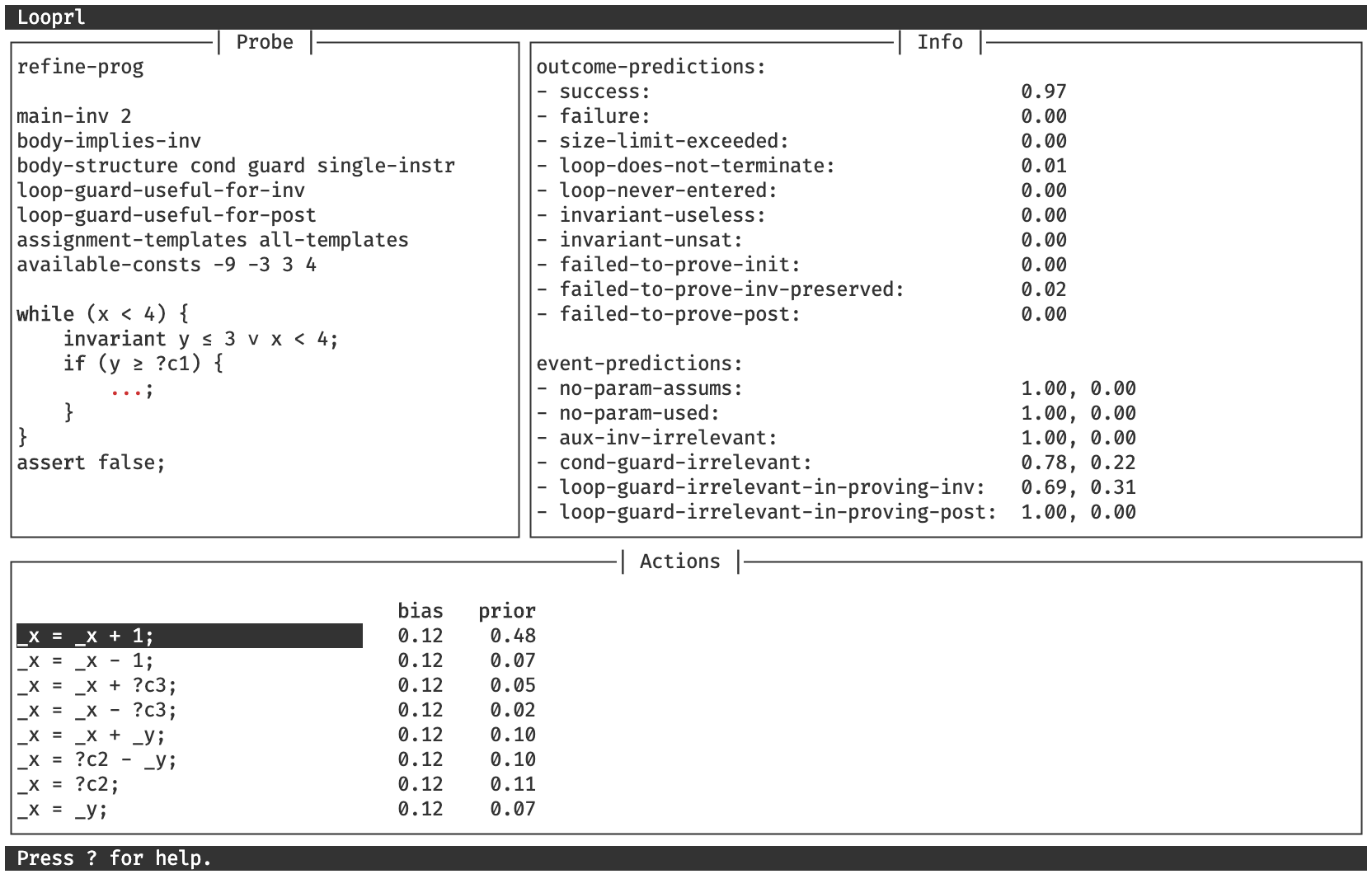}
\caption{Visualizing the teacher strategy with the \looprl{} UI. This screenshot captures a choice point where the network is tasked with adding an assignment to the true branch of the conditional within the loop body. Several templates are proposed, which are going to be refined in turn. The upper left pane (i.e. the \emph{Probe} pane) features all contextual information that is sent to the network to help it make a choice. Among this information, we can see a list of all constraints that the generated problem should ideally satisfy. The \emph{Info} pane gives us some insights into the network's prediction about future events and outcome. For example, the network estimates with $0.97$ probability that a valid problem will be generated (along with a $0.02$ probability that a failure will be encountered due to the invariant not being preserved by the loop body). The network also estimates a $31\%$ risk that the generated problem will violate the soft constraint according to which the loop guard should be relevant for proving the invariant.}\label{fig:teacher-ap-ui}
\end{figure}

We show examples of using the \looprl{} user interface to inspect the solver and teacher strategies in Figures~\ref{fig:solver-ap-ui} and~\ref{fig:teacher-ap-ui} respectively.

\section{Implementing Abduction}\label{ap:abduction}

Both the teacher and solver strategies we use as examples in this paper rely on an \texttt{abduct} function that takes as an input a formula $F$ and then either proves it valid or returns a (possibly empty) set of assumptions $A$ such that $A \limply F$ can be proved to be valid.

Implementing such a function is hard in the general case and so an implementation of \texttt{abduct} may leverage nondeterminism. However, because we are only dealing with arithmetic of limited complexity in this work, we implemented a fully-specified abduction procedure for linear integer arithmetic that relies on Fourier-Motzkin elimination \cite{DBLP:journals/jct/DantzigE73}.

When given a formula $F$, our abduction procedure first rewrites it in conjunctive normal form as $F = \bigwedge_{i=1}^n \bigvee_j F_{ij}$ where $F_{ij}$ are atomic formulas of the form $\sum_k a_kx_k \odot c$ where $\odot \in \{\ge, =\}$ and $c, a_k$ are integer constants.

\subsection{Case where $n = 1$}

If $F$ can be expressed as a disjunction of atomic formulas, we can compute abduction suggestions as follows:
\begin{inparaenum}[\it i)]
  \item one considers the negation of $F$, obtaining a set of atomic assumptions and
  \item one derives as many consequences as possible from those assumptions using Fourier-Motzkin elimination (linearly combining inequalities so as to eliminate variables).
\end{inparaenum}
If a contradiction is derived, then $F$ is valid. If no contradiction is dervied given some timeout, then the negation of any derived consequence can be considered as an abduction candidate.

\paragraph{Example} Suppose we want to compute abduction candidates for $F = x \ge 0 \limply x+y \geq 1$. The conjunctive normal form of $F$ is $(x < 0 \,\vee\, x + y \geq 1)$. Taking the negation yields $(x \ge 0 \,\wedge\, x+y<1)$, which we normalize into the set of assumptions $\{x \ge 0,  -x-y \ge 0\}$ (all variables are integers). Then, we can take a Fourier-Motzkin step by adding these two assumptions and derive the following consequence: $-y \ge 0$. After this, no other reasoning step is applicable end we end up with the following set of facts: $\{x \ge 0,  -x-y \ge 0, -y \ge 0\}$. We therefore suggest the following abduction candidates: $x < 0$, $x + y > 0$ and $y > 0$.

\subsection{Case where $n > 1$}

If the conjunctive normal form of $F$ has two conjuncts or more, we apply the procedure above on each conjunct separately. For example, suppose that $F = G \wedge H$, $G$ admits a set of abduction candidates $\{A_i\}_i$ and $H$ admits a set of abduction candidates $\{B_i\}_i$. One possibility would be to return  all $\{A_i \wedge B_j\}_{ij}$ combinations as abduction candidates for $F$. However, doing so can quickly result in a combinatorial explosion. Therefore, our implementation does something different and returns the union of the $\{A_i\}_i$ and $\{B_i\}_i$ instead. Doing so, it cannot provide the guarantee that any resulting abduction candidate $A$ is sufficient in implying $F$. Rather, abduction candidates are seen as suggestions to unblock one part of the proof (i.e. enable proving one conjunct of $F$) but not necessarily the whole proof.


\section{Training Hyperparameters}\label{ap:hyperparams}

We provide an exhaustive list of all hyperparameter values used in our experiments in Table~\ref{tab:hyperparams}.

{
\small
\renewcommand{\arraystretch}{1.5}
\begin{longtable}{lcl}
\expandableinput generated/hyperparams.tex
\caption{Training hyperparameters for the experiments described in Section~\ref{sec:experiments}. Hyperparameters are organized according to a hierarchichal structure that is represented using indentation.}
\label{tab:hyperparams}
\end{longtable}
}


\section{Network Architecture and Choice Points Encoding}\label{ap:network}

\subsection{Network Architecture}

Neural networks that are used as oracles for nondeterministic strategies take as an input a \emph{choice point} and return
\begin{inparaenum}[\it i)]
  \item a probability distribution over all available choices,
  \item some success and failure probability estimates and
  \item some event occurence probability estimates (see Section~\ref{sec:vec-rewards}).
\end{inparaenum}
In turn, a {choice point} is represented as
\begin{inparaenum}[\it i)]
  \item a \emph{probe}~\cite{selsambeyond} that encodes all relevant state information provided to \texttt{choose} (see Figure~\ref{fig:solver-code} for examples) and
  \item a list of possible choices that were passed as arguments to \texttt{choose}.
\end{inparaenum}

Our proposed network architecture works as follows. Given a choice point, the probe is encoded using a Dynamic Graph Transformer (DGT) neural network~\cite{selsam2020universal}. DGT networks are similar to Transformers~\cite{DBLP:conf/nips/VaswaniSPUJGKP17} but they allow leveraging some additional graph structure over the source tokens by associating edge types to learned attention biases. Each choice is encoded separately using another DGT. It is then concatenated with the probe encoding and passed to a combiner network that outputs a score. Scores are normalized into a probability distribution using a softmax operation. The probe encoding is also passed to a value head that produces outcome and event predictions. Batches of choice points can be evaluated efficiently in parallel using scatter operations~\cite{Fey/Lenssen/2019}.


\subsection{Encoding Programs and Formulas}

All data that is to be passed to the neural network must be encodable into a sequence of tokens with an optional graph structure. This is the case of programs and formulas in particular. We use standard techniques to encode those~\cite{DBLP:conf/iclr/HellendoornSSMB20,DBLP:conf/nips/ShivQ19}:

\begin{itemize}
  \item Abstract syntax trees (ASTs) are encoded into a sequence of tokens in polish notation order (e.g. $x + 3y$ is encoded as ``\texttt{PLUS VAR(x) MUL CONST(3) VAR(y)}''). The edges in the original AST are preserved as graph edges to be passed to the DGT network.
  \item We use a different positional encoding scheme for tokens that leverages the tree structure of the underlying AST~\cite{DBLP:conf/nips/ShivQ19}. Doing so has been demonstrated to result in better generalization capabilities~\cite{DBLP:conf/iclr/HahnSKRF21}.
  \item Identifier names are randomly mapped into a finite number of unique ids for each choice point. Unique ids are encoded using a one-hot encoding scheme. If a choice point introduces more names than there are unique ids available (rare), the last occuring names are made to share a single uid that is flagged to indicate a naming conflict.
  \item Numerical constants with an absolute value greater than $3$ are represented with generic \texttt{POS\_CONST} and \texttt{NEG\_CONST} tokens. A binary encoding of their value is also provided to the network. Moreover, special edges are added to compare all numerical constants involved in a program or formula (see Table~\ref{tab:edges}).
  \item Following~\cite{DBLP:conf/iclr/HellendoornSSMB20}, we also add semantic edges to the encoding of programs and formulas that reflect the way information flows within them. Details can be consulted in Table~\ref{tab:edges}.
\end{itemize}

\begin{table}
  \begin{center}
    \begin{tabularx}{\textwidth}{p{3cm}X}
  \toprule
  \textbf{Edge type} & \textbf{Description} \\
  \midrule
  \texttt{PARENT} & Connect any token to its parent in the syntax tree.  \\
  \texttt{PREV\_SIBLING} & Connect any token to its previous sibling in the syntax tree. \\
  \texttt{PREV\_LEXICAL\_USE} & Connect any token associated with name $s$ to the previous occurence of $s$. \\
  \midrule
  \texttt{LAST\_READ} & Connect a variable occurence to places where it was possibly read last. \\
  \texttt{LAST\_WRITE} & Connect a variable occ. to places where it was possibly written last. \\
  \texttt{GUARDED\_BY} & Connect a variable occ. to assumptions made about it. \\
  \texttt{GUARDED\_BY\_NEG} & Connect a variable occ. to negated assumptions made about it. \\
  \texttt{COMPUTED\_FROM} & Connect the lhs of any assignment to the variables in its rhs. \\
  \midrule
  \texttt{SAME\_CONST} & Connect two identical numerical constants. \\
  \texttt{SMALLER\_CONST} & Compare two different numerical constants. \\
  \bottomrule
\end{tabularx}
  \end{center}
  \medskip
  \caption{Edge types used to encode formulas and programs. We organize these edges into three groups: syntactic edges, semantic edges and numerical edges. Within a conditional statement, every variable in the \texttt{if} branch is connected to the guard using a \texttt{GUARDED\_BY} edge whereas every variable in the \texttt{else} branch is connected to the guard using a \texttt{GUARDED\_BY\_NEG} edge.}\label{tab:edges}
\end{table}


\end{document}